\documentclass[sigconf]{acmart}

\renewcommand\footnotetextcopyrightpermission[1]{} 
\AtBeginDocument{%
  \providecommand\BibTeX{{%
    \normalfont B\kern-0.5em{\scshape i\kern-0.25em b}\kern-0.8em\TeX}}}

\setcopyright{none}

\usepackage{fancyhdr}
\usepackage{hyperref}
\usepackage{algorithm}
\usepackage{algorithmic}
\usepackage{xspace}
\usepackage[font=small,labelfont=bf]{caption}
\usepackage[caption=false]{subfig}
\usepackage{amsmath}
\usepackage{mathtools}
\usepackage{xcolor}
\usepackage{amsfonts}
\usepackage{amssymb}
\usepackage{pifont}
\usepackage{color}
\xspaceaddexceptions{]\}}
\newcommand{\ignore}[1]{}

\newcommand{\old}[1]{}
\newcommand{\fig}[1]{Figure~\ref{#1}}
\newcommand{\sect}[1]{Section~\ref{#1}}
\newcommand{\tab}[1]{Table~\ref{#1}}

\newcommand{\tdimm}[0]{\texttt{TensorDIMM}\xspace}
\newcommand{\pmem}[0]{\texttt{PMEM}\xspace}
\newcommand{\tdm}[0]{\texttt{TDIMM}\xspace}

\newcommand{\cpuonly}[0]{\texttt{CPU-only}\xspace}

\newcommand{\cpugpu}[0]{\texttt{CPU-GPU}\xspace}
\newcommand{\gpuonly}[0]{\texttt{GPU-only}\xspace}

\newcommand{\tdimms}[0]{\texttt{TensorDIMM}s\xspace}

\newcommand{\tisa}[0]{\texttt{TensorISA}\xspace}
\newcommand{\tnode}[0]{\texttt{TensorNode}\xspace}
\newcommand{\lookup}[0]{\texttt{GATHER}\xspace}
\newcommand{\reduce}[0]{\texttt{REDUCE}\xspace}
\newcommand{\average}[0]{\texttt{AVERAGE}\xspace}

\newcommand\blfootnote[1]{%
\begingroup
\renewcommand\thefootnote{}\footnote{#1}%
\addtocounter{footnote}{-1}%
\endgroup
}

\begin{document}

\title[TensorDIMM: A Near-Memory Processing Architecture for Sparse Embedding Layers in Deep Learning]{TensorDIMM: A Practical Near-Memory Processing Architecture \\for Embeddings and Tensor Operations in Deep Learning}

\author{Youngeun Kwon}
\affiliation{%
  \institution{School of Electrical Engineering\\KAIST}}
\email{yekwon@kaist.ac.kr}

\author{Yunjae Lee}
\affiliation{%
  \institution{School of Electrical Engineering\\KAIST}}
\email{yunjae408@kaist.ac.kr}

\author{Minsoo Rhu}
\affiliation{%
  \institution{School of Electrical Engineering\\KAIST}}
\email{mrhu@kaist.ac.kr}

\renewcommand{\shortauthors}{Youngeun Kwon, Yunjae Lee, Minsoo Rhu}


\begin{abstract}

	Recent studies from several hyperscalars pinpoint to embedding layers as the
	most memory-intensive deep learning (DL) algorithm being deployed in today's
	datacenters. This paper addresses the memory capacity and bandwidth
	challenges of embedding layers and the associated tensor operations.  We
	present our vertically integrated hardware/software co-design, which includes a custom DIMM module
	enhanced with near-memory processing cores tailored for DL tensor operations.
	These custom DIMMs are populated inside a GPU-centric system interconnect as
	a remote memory pool, allowing GPUs to utilize for scalable memory bandwidth
	and capacity expansion. A prototype implementation of our proposal on real DL
	systems shows an average $6.2$$-$$17.6\times$ performance improvement on state-of-the-art
		DNN-based recommender systems.

	\end{abstract}

\maketitle

\blfootnote{
This is the author preprint version of the work. The authoritative version will appear in the Proceedings of the $52^{\text{nd}}$ Annual IEEE/ACM International Symposium on Microarchitecture (MICRO-52), October 12--16, 2019, Columbus, OH, USA.
}
\section {Introduction}
\label{sect:intro}

\begin{figure}[t!] \centering
\vspace{0.3em}
\hspace{3em}\includegraphics[width=0.40\textwidth]{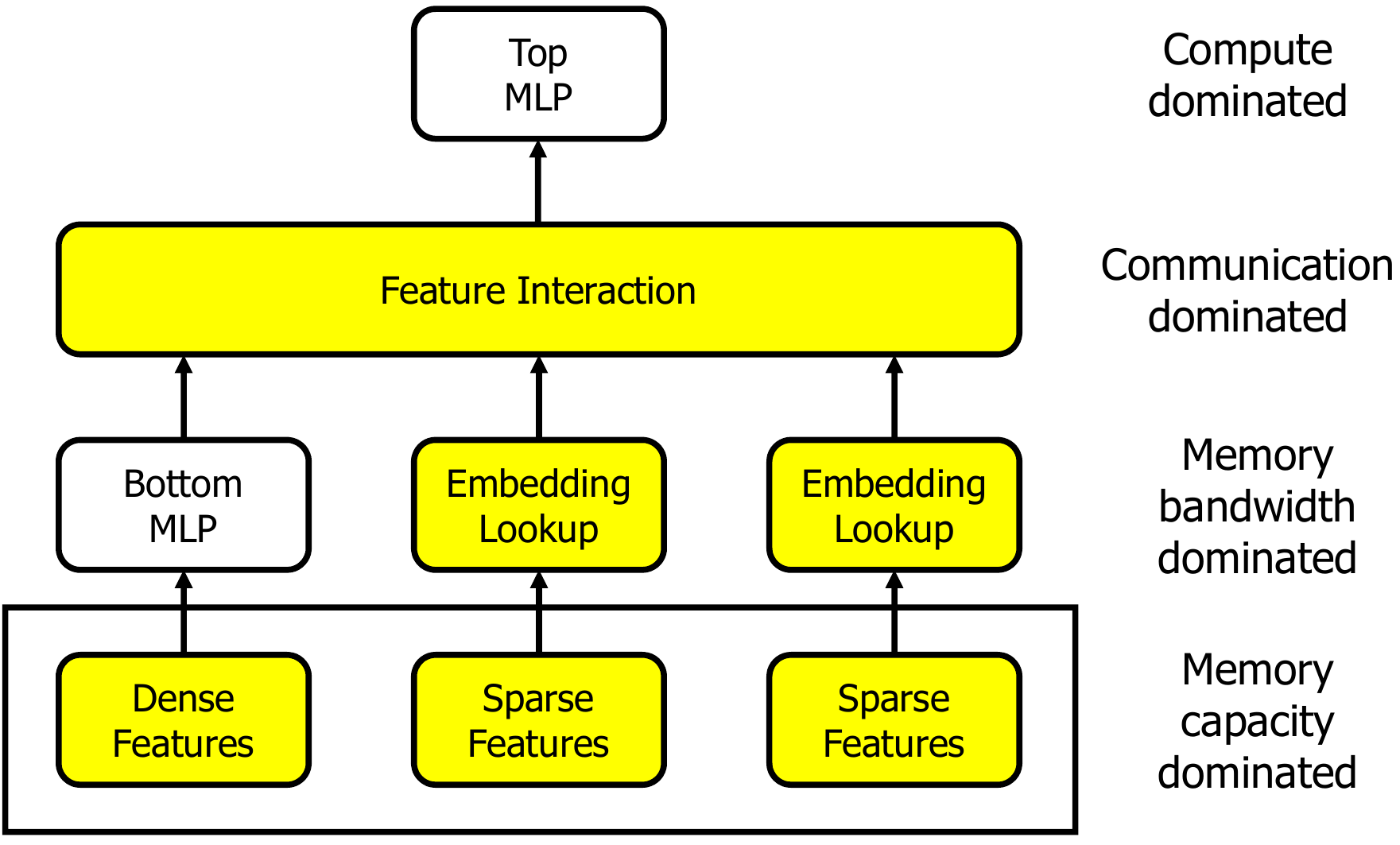}
\caption{
Topological structure of emerging DL applications. The figure is reproduced
	from Facebook's keynote speech at the Open Compute Project summit
	2019~\cite{ocp_speech}, which calls out for architectural solutions
	providing ``High memory bandwidth and capacity
	for embeddings''. This paper specifically addresses this important
			memory wall problem in emerging DL applications: i.e., the \emph{non}-MLP portions (yellow) in this figure.
}
\vspace{-0.7em}
\label{fig:upcoming_dnns}
\end{figure}

Machine learning (ML) algorithms based on deep neural networks (DNNs), also
known as deep learning (DL), is scaling up rapidly. To satisfy the computation
needs of DL practitioners, GPUs or custom-designed accelerators for DNNs, also
known as neural processing units (NPUs) are widely being deployed for
accelerating DL workloads.  Despite prior studies on enhancing the compute
throughput of
GPUs/NPUs~\cite{mcm_gpu,tpu1,eyeriss,cnvlutin,scnn,dadiannao,diannao,stripes},
a research theme that has received relatively less attention is how computer
system architects should go about tackling the ``memory wall'' problem in DL:
emerging DL algorithms demand both high memory capacity and bandwidth, limiting
the types of applications that can be deployed under practical constraints.  In
particular, recent studies from several
hyperscalars~\cite{park:2018:fb,hestness:2019:ppopp,dean:2018:goldenage,facebook_dlrm,dlrm:arch}
pinpoint to \emph{embedding lookups} and \emph{tensor manipulations} (aka
\emph{embedding layers}, \sect{sect:embedding}) as the most memory-intensive
algorithm deployed in today's datacenters, already reaching several hundreds of
GBs of memory footprint, even for inference.  Common wisdom in conventional DL
workloads (e.g., convolutional and recurrent neural networks, CNNs and RNNs)
was that convolutions or matrix-multiplications account for the majority of
inference time~\cite{scnn,song:2015:eie}. However, emerging DL applications
employing embedding layers exhibit drastically different characteristics as the
embedding lookups and tensor manipulations (i.e., \emph{feature interaction} in
\fig{fig:upcoming_dnns}) account for a significant fraction of execution time.
Facebook for instance states that embedding layers take up $34\%$ of the
execution time of all DL workloads deployed in their
datacenters~\cite{park:2018:fb}.

Given this landscape, this paper focuses on addressing the memory capacity and
bandwidth challenges of embedding layers (\fig{fig:upcoming_dnns}).  Specifically, we focus our attention on
\emph{recommender systems}~\cite{recsys} using embeddings which is one of the
most common DL workloads deployed in today's datacenters for numerous
application domains like advertisements, movie/music recommendations, news feed, and etc.  As
detailed in \sect{sect:embedding}, the model size of embedding
layers (typically several hundreds of GBs~\cite{hestness:2019:ppopp,park:2018:fb}) far
exceeds the memory capacity of GPUs. As a result, the solution vendors
take is to store the entire embedding lookup table inside the
capacity-optimized, low-bandwidth CPU memory and deploy the application
1) by only using CPUs for the entire computation of DNN inference, or 2) by employing a hybrid CPU-GPU approach
where the embedding lookups are conducted on the CPU but the rest are handled on the GPU.
We observe that both these approaches leave significant performance left on the table,
experiencing an average $7.3$$-$$20.9\times$
slowdown compared to a hypothetical, GPU-only version which
assumes the entire embeddings can be stored in the GPU memory
(\sect{sect:recsys_bottleneck}).  Through a detailed application
characterization study, we root-cause the reason behind such performance
loss to the following factors.  First, the embedding vectors are read
out using the low-bandwidth CPU memory, incurring significant latency overheads
compared to when the embeddings are read out using the bandwidth-optimized GPU memory.
Second, the low computation throughput of CPUs can significantly slowdown the computation-heavy DNN
execution step when solely relying on CPUs, whereas the hybrid CPU-GPU version can suffer from the
latencies in copying the embeddings from CPU to GPU
memory over the thin PCIe channel.

	To tackle these challenges, we present a vertically integrated,
	hardware/software co-design that fundamentally addresses the memory (capacity
			and bandwidth) wall problem of embedding layers. Our proposal encompasses
	multiple levels in the hardware and software stack as detailed
	below.

{\bf (Micro)architecture.} We present \tdimm, which is based on commodity
buffered DIMMs but further enhanced with near-memory processing
(NMP) units customized for key DL tensor operations, such as embedding
\emph{gathers} and \emph{reductions}. 
The NMP units in \tdimm are designed to conduct embedding gathers and reduction operations
``near-memory'' which drastically reduces the latency in fetching the embedding vectors
and reducing them,
providing significant improvements in effective
communication bandwidth and performance.  Additionally, our \tdimm leverages
commodity DRAM devices as-is, so another key advantage of our proposal as opposed to
prior in-memory architectures~\cite{ISAAC,valavi:cnn,tu2018rana} is its practicality and ease of
implementation.

{\bf ISA extension and runtime system.} Building on top of our \tdimm architecture,
	we propose a custom tensor ISA and runtime system that provides
	\emph{scalable} memory bandwidth and capacity expansion for embedding layers.
	Our proposal entails 1) a carefully designed ISA tailored
	for DL tensor operations (\tisa), 2) an efficient address mapping scheme for
	embeddings, and 3) a runtime system that effectively
	utilizes \tdimms for tensor operations.  The \tisa has been
	designed from the ground-up to conduct key DL tensor operations in a memory
	bandwidth efficient manner.  An important challenge with conventional memory
	systems is that, regardless of how many DIMMs/ranks are physically available per
	each memory channel, the maximum memory bandwidth provided to the memory
	controller is fixed.  \tisa has been carefully co-designed with both \tdimm
	and the address mapping scheme so that the aggregate memory bandwidth
	provided to our NMP units increases proportional to the number of \tdimms.
	In effect, our proposal offers a platform for \emph{scalable} memory bandwidth
	expansion for embedding layers. Compared to
	the baseline system, our default \tdimm configuration offers
	an average $4\times$ increase in memory bandwidth for key DL tensor operations.

{\bf System architecture.} Our final proposition is to aggregate a pool of
\tdimm modules as a \emph{disaggregated memory node} (henceforth referred to as
		\tnode) in order to provide scalable memory capacity expansion. A key to
our proposal is that \tnode is interfaced inside the NVLINK-compatible,
		GPU-side high-bandwidth system interconnect\footnote{While we demonstrate 
			\tdimm's merits under the context of a high-bandwidth GPU-side system interconnect, the effectiveness of our proposal remains intact
			for NPUs (e.g., Facebook's Zion interconnect~\cite{zion}).
		}.  In state-of-the-art DL systems, the
		GPUs are connected to a high-bandwidth switch such as
		NVSwitch~\cite{nvswitch} which allows high-bandwidth, low-latency data
		transfers between any pair of GPUs.  Our proposal utilizes a pooled memory
		architecture ``inside'' the high-bandwidth GPU-side interconnect, which is
		fully populated with capacity-optimized memory DIMMs -- but in our case,
		the \tdimms.  The benefits of interfacing \tnode within the GPU
		interconnect is clear: by storing the embedding lookup tables inside the
		\tnode, GPUs can copy in/out the embeddings much faster than the
		conventional CPU-GPU based approaches (i.e., approximately
				$9\times$ faster than PCIe, assuming NVLINK(v2)~\cite{nvlink}).
		Furthermore, coupled with the memory bandwidth amplification effects of
		\tisa, our \tnode offers scalable expansion of ``both'' memory capacity and
		bandwidth.  Overall, our vertically integrated solution provides an average
		$6.2$$-$$15.0\times$ and $8.9$$-$$17.6\times$ speedup in inference time compared
		to the CPU-only and hybrid CPU-GPU implementation of recommender systems, respectively.
		\noindent To summarize our {\bf key contributions}:

\begin{itemize}

\item The computer architecture community has so far primarily focused on
accelerating the computationally intensive, ``dense'' DNN layers (e.g., CNNs/RNNs/MLPs).
To the best of our knowledge, this work is the first to identify, analyze, and explore architectural
solutions for ``sparse'' embedding layers, a highly important building block with significant industrial importance
for emerging DL applications.

\item We propose \tdimm, a practical near-memory processing architecture built on top
of commodity DRAMs which offers scalable increase in
both memory capacity and bandwidth for embedding gathers and tensor operations.

\item We present \tnode, a \tdimm-based disaggregated memory system for DL. The efficiency of our solution
is demonstrated as a proof-of-concept software prototype on
a high-end GPU system, achieving significant performance improvements
than conventional approaches.

\end{itemize}

\section{Background} 
\label{sect:background}

\subsection{Buffered DRAM Modules}
\label{sect:buffered_dimm}

In order to balance memory capacity and bandwidth, commodity DRAM devices that
are utilized in unison compose a \emph{rank}. One or more ranks are packaged
into a memory module, the most popular form factor being the dual-inline memory
module (DIMM) which has $64$ data I/O (DQ) pins. Because a memory channel is
typically connected to multiple DIMMs, high-end CPU memory controllers often
need to drive hundreds of DRAM devices in order to deliver the command/address
(C/A) signals through the memory channel. Because modern DRAMs operate in the GHz
range, having hundreds of DRAM devices be driven by handful of memory controllers
leads to signal integrity issues. Consequently, server-class DIMMs typically employ
a \emph{buffer device} per each DIMM (e.g., registered DIMM~\cite{rdimm_8gb} or load-reduced DIMM~\cite{lrdimm})
which is used to repeat the C/A signals to reduce the high capacitive load and resolve 
signal integrity issues.
	 Several prior work from both industry~\cite{centaur}
and in academia~\cite{nda1,chameleon,mcn} have explored the possibility of utilizing
this buffer device space to add custom logic designs to address specific application needs.
IBM's Centaur DIMM~\cite{centaur} for instance utilizes a buffer device to add
a $16$ MB eDRAM L4 cache and a custom interface between DDR PHY and IBM's proprietary memory interface.

\subsection{System Architectures for DL}
\label{sect:npu_arch}

As the complexity of DL applications skyrocket, there has been a growing trend
towards dense, scaled-up system node design with multiple PCIe-attached
co-processor devices (i.e., DL accelerators such as
		GPUs/NPUs~\cite{volta_v100,tpu2}) to address the problem size growth.  A
multi-accelerator device solution typically works on the same problem in
parallel with occasional inter-device communication to share intermediate
data~\cite{alex_weird_trick}.  Because such inter-device communication often
lies on the critical path of parallelized DL applications, system vendors are
employing high-bandwidth interconnection fabrics that
utilize custom high-bandwidth signaling links (e.g., NVIDIA's DGX-2~\cite{dgx_2} or
		Facebook's Zion system interconnect fabric~\cite{zion}).  NVIDIA's
DGX-2~\cite{dgx_2} for instance contains $16$ GPUs, all of which are
interconnected using a NVLINK-compatible high-radix (crossbar) switch called
NVSwitch~\cite{nvswitch}.  NVLINK provides $25$ GB/sec of full-duplex
uni-directional bandwidth per link~\cite{nvlink}, so any given GPU within DGX-2
can communicate with any other GPU at the full uni-directional bandwidth up to
$150$ GB/sec via NVSwitch. Compared to the thin uni-directional bandwidth of
$16$ GB/sec (x16) under the CPU-GPU PCIe(v3) bus, such high-bandwidth GPU-side
interconnect enables an order of magnitude faster data transfers.

\subsection{DL Applications with Embeddings}
\label{sect:embedding}

{\bf DNN-based recommender systems.} Conventional DL applications for inference (e.g.,
		CNNs/RNNs) generally share a common property where its overall memory
footprint fits within the (tens of GBs of) GPU/NPU physical memory. However, recent studies from several hyperscalars~\cite{park:2018:fb,hestness:2019:ppopp,dean:2018:goldenage,facebook_dlrm,dlrm:arch} call out for imminent
system-level challenges in emerging DL workloads that are extremely memory
(capacity and bandwidth) limited.  Specifically, these hyperscalars pinpoint to
\emph{embedding layers} as the most memory-intensive algorithm deployed in
their datacenters.  One of the most  widely deployed DL
application using embeddings is the \emph{recommender system}~\cite{recsys}, which is
used in numerous application domains such as advertisements (Amazon, Google,
		Ebay), social networking service (Facebook, Instagram), movie/music/image
recommendations (YouTube, Spotify, Fox, Pinterest), news feed (LinkedIn), and
many others.  Recommendation is typically formulated as a problem of predicting
the probability of a certain event (e.g., the probability of a Facebook user
clicking ``like'' for a particular post), where a ML model estimates the likelihood of one
or more events happening at the same time. Events or items with the highest
probability are ranked higher and recommended to the user.

Without going into a comprehensive review of numerous prior literature on recommendation models, 
				we emphasize that current state-of-the-art recommender
systems have evolved into utilizing (not surprisingly) DNNs.  While there
exists variations regarding how the DNNs are constructed, a commonly
employed topological structure for DNN-based recommender systems is the neural network
based collaborative filtering algorithm~\cite{ncf}.  Further advances led
to the development of more complex models with wider and deeper vector
dimensions, successfully being applied and deployed in commercial
					 user-facing products~\cite{park:2018:fb}.

\begin{figure}[t!] \centering
\vspace{0.5em}
\includegraphics[width=0.49\textwidth]{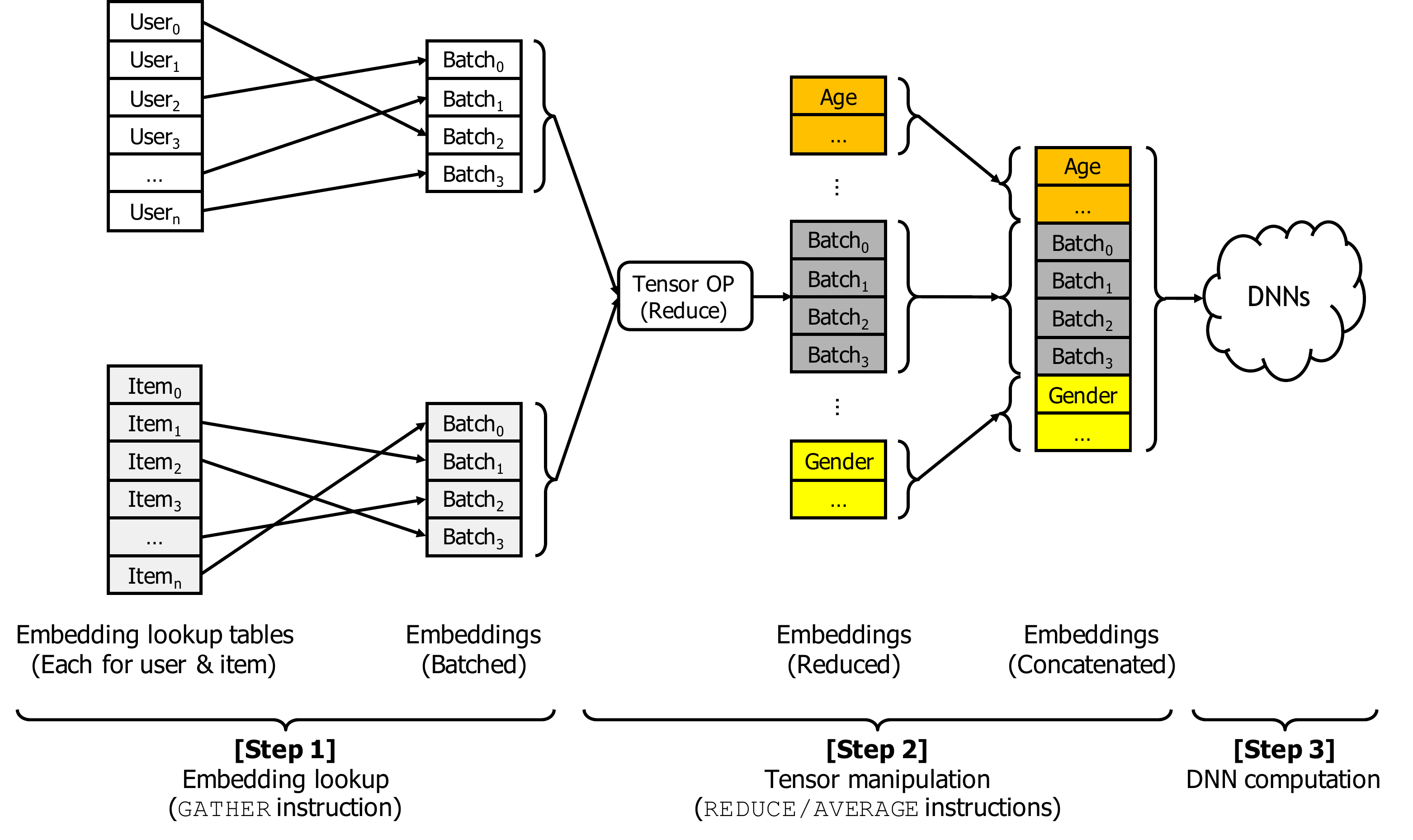}
\vspace{-1.5em}
\caption{
A DNN based recommender system. The embedding layer typically consists of the
	following two steps. (1) The embedding lookup stage where embeddings are ``\emph{gathered}'' from
	the (potentially multiple, two tables in this example) look-up tables  up to its batch size and form
	(batched) embedding tensors. These tensors go through several tensor manipulation operations
	to form the final embedding tensor to be fed into the DNNs.
	Our proposal utilizes custom tensor ISA extensions (\texttt{GATHER}/\texttt{REDUCE}/\texttt{AVERAGE}) to accelerate this process,
	which we detail in \sect{sect:tensor_isa}.
}
\vspace{-0.5em}
\label{fig:ncf}
\end{figure}

{\bf Embedding lookups and tensor manipulation.} \fig{fig:ncf} illustrates the usage of embeddings
in recommender systems that incorporate DNNs~\cite{ncf}.  The inputs
to the DNNs (which are typically constructed using fully-connected layers or
		multi-layer perceptrons, FCs or MLPs) are constructed as a combination of
dense and sparse features. Dense features are commonly represented as a
vector of real numbers whereas sparse features are initially represented as
indices of one-hot encoded vectors. These one-hot indices are used to query the
\emph{embedding lookup table} to project the sparse indices into a dense vector
dimension space. The contents stored in these lookup tables are called
\emph{embeddings}, which are trained to extract deep learning features (e.g.,
		Facebook trains embeddings to extract information regarding what pages a
		particular user liked, which are utilized to recommend relevant contents or
		posts to the user~\cite{park:2018:fb}). The embeddings read out of the lookup table are
combined with other dense embeddings (\emph{feature interaction} in \fig{fig:upcoming_dnns})
	from other lookup tables using tensor concatenation or \emph{tensor ``reductions''}
such as element-wise additions/multiplications/averages/etc
to generate an output tensor, which is then forwarded to the
DNNs to derive the final event probability (\fig{fig:ncf}).

{\bf Memory capacity limits of embedding layers.} A key reason why embedding
layers consume significant memory capacity is because each user (or each item)
	requires a unique embedding vector inside the lookup table. The total number
	of embeddings therefore scales proportional to the number of users/items,
	rendering the overall memory footprint to exceed several hundreds of GBs just
	to keep the model weights themselves, even for inference.  Despite its high
	memory requirements, embedding layers are favored in DL applications because
	it helps improve the model quality: given enough memory capacity, users seek
	to further increase the model size of these embedding layers using \emph{larger
	embedding dimensions} or by using \emph{multiple embedding tables} to combine
	multiple dense features using tensor reduction operations.

\section{Motivation}
\label{sect:motivation}

\subsection{Memory (Capacity) Scaling Challenges} 
\label{sect:memory_scaling_challenge}

	High-end GPUs or NPUs commonly employ a bandwidth-optimized,
	on-package 3D stacked memory (e.g., HBM~\cite{hbm} or HMC~\cite{hmc}) in
	order to deliver the highest possible memory bandwidth to the on-chip
	compute units. Compared to capacity-optimized DDRx most commonly adopted in CPU
	servers, the bandwidth-optimized stacked memory is capacity-limited,
	only available with several tens of GBs of storage. While one might expect
	future solutions in this line of product to benefit from higher memory density,
	there are technological constraints and challenges in increasing the capacity
	of these 3D stacked memory in a \emph{scalable} manner. First, stacking more DRAM
	layers vertically is constrained by chip pinout required to drive the added
	DRAM stacks, its wireability on top of silicon interposers, and thermal
	constraints.  Second, current generation of GPUs/NPUs are already close to
	the reticle limits of processor die size\footnote{NVIDIA V100 for instance has already
		reached the reticle limits of $815$ mm$^2$ die area, forcing researchers to
			explore alternative options such as multi-chip-module
			solutions~\cite{mcm_gpu} 	to continue computational scaling.
	
	} so adding
			more 3D stack modules within a package inevitably leads to
			sacrificing area budget for compute units.

\subsection{Memory Limits in Recommender System}
\label{sect:recsys_bottleneck}

As discussed in \sect{sect:embedding}, the model size of embedding lookup
tables is in the order of several hundreds of GBs, far exceeding the memory
capacity limits of GPUs/NPUs. Due to the memory scaling limits of on-package
stacked DRAMs, a solution vendors take today is to first store the embedding
lookup tables in the CPU and read out the
embeddings using the (capacity-optimized but bandwidth-limited) CPU memory. Two possible
implementations beyond this step are as follows. The CPU-only version (\cpuonly)
goes through the rest of the inference process using the CPU without relying
upon the GPU. A \emph{hybrid} CPU-GPU approach (\cpugpu)~\cite{nvidia_blog_recsys} on the
other hand copies the embeddings to the GPU memory over PCIe using
\texttt{cudaMemcpy}, and once the CPU$\rightarrow$GPU data transfer is
complete, the GPU initiates various tensor manipulation operations to form the
input tensors to the DNN, followed by the actual DNN computation step
(\fig{fig:ncf}).

\begin{figure}[t!] \centering
	 \vspace{1em}
	\includegraphics[width=0.475\textwidth]{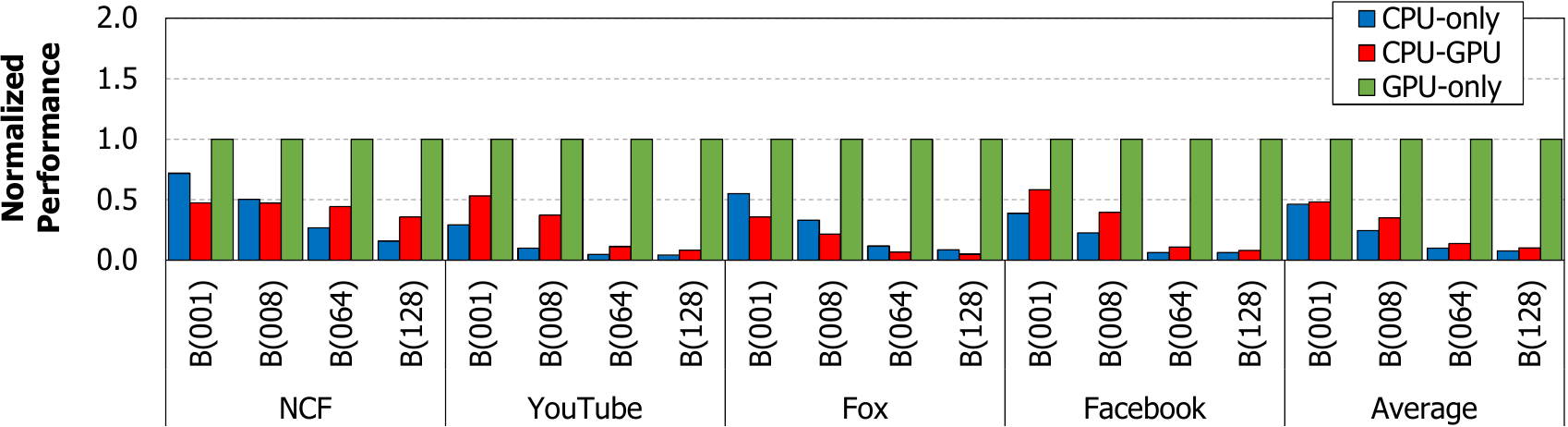}
\caption{
	Performance of baseline \cpuonly and hybrid \cpugpu versions of recommender
		system, normalized to an oracular \gpuonly version. \texttt{B(N)}
	represents an inference with batch size \texttt{N}. \cpuonly
		exhibits some performance advantage than the \cpugpu version for certain low batch inference
		scenarios, but both \cpuonly and \cpugpu suffers from significant performance loss compared to oracular \gpuonly.
		\sect{sect:methodology}
	details our evaluation methodology.
}
\vspace{-0.2em}
\label{fig:perf_hybrid_vs_cpu_only}
\end{figure}

Given this landscape, DL practitioners as well as system designers are faced
with a conundrum when trying to deploy recommender systems for inference.  From
a DL algorithm developer's perspective, you seek to add more embeddings
(i.e., more embedding lookup tables to combine various embedding vectors using
 tensor manipulations, such as tensor reduction) and increase embedding
dimensions (i.e., larger embeddings) for complex feature interactions as it improves model quality.
Unfortunately, fulfilling the needs of these algorithm developers bloats up overall
memory usage (\fig{fig:embedding_mem_usage})
and inevitably results in resorting to the capacity-optimized CPU
		memory to store the embedding lookup tables.  Through a detailed characterization study,
		we root-cause the following three factors as {\bf key limiters} of prior
		approaches (\fig{fig:perf_hybrid_vs_cpu_only}) relying on CPU memory for storing embeddings:

\begin{figure}[t!] \centering
\includegraphics[width=0.50\textwidth]{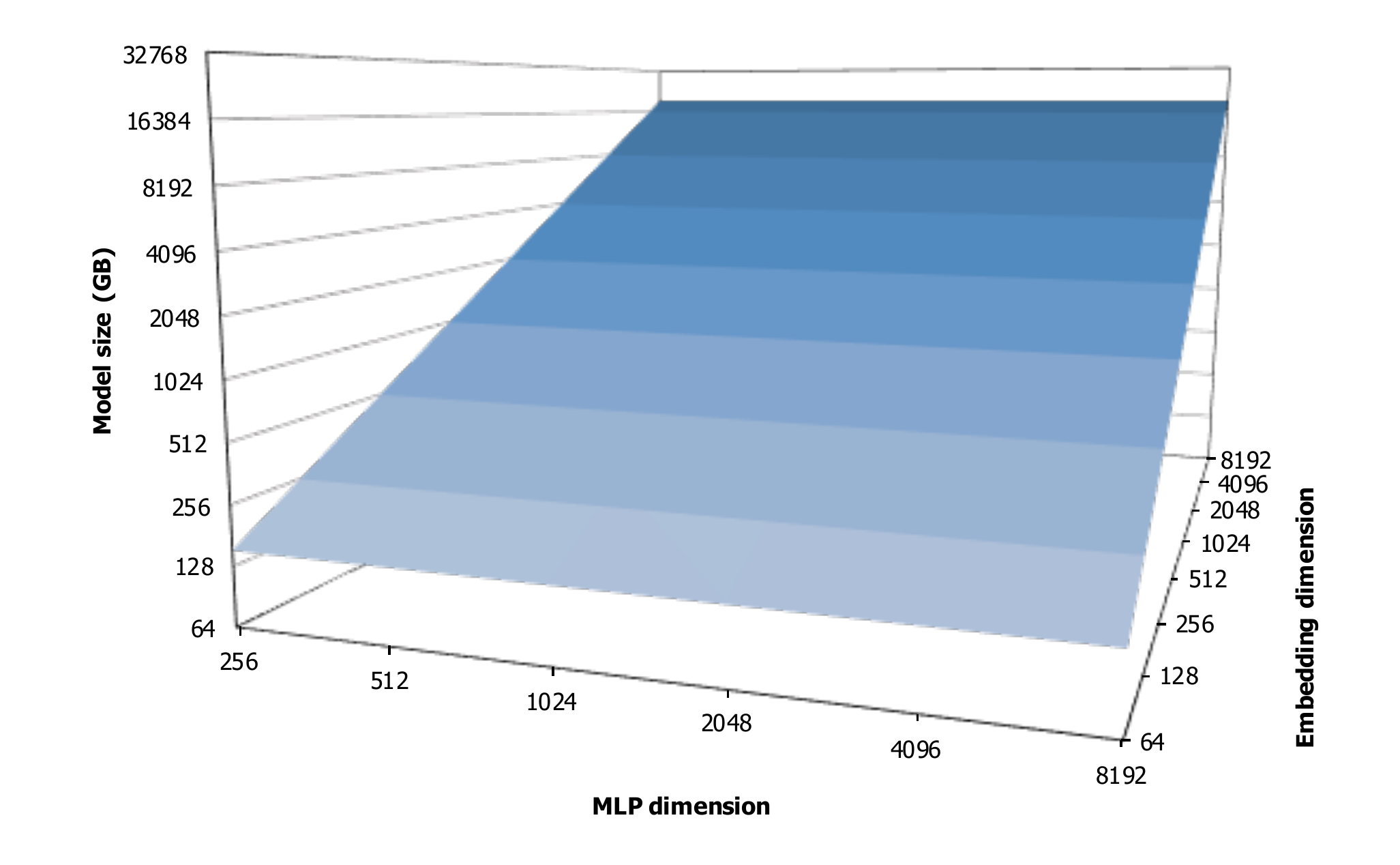}
\vspace{-2.0em}
\caption{
Model size growth of neural collaborative filtering (NCF~\cite{ncf}) based recommender system
when the MLP layer dimension size (x-axis) and the embedding vector dimension size (y-axis)
is scaled up. Experiment assumes the embedding lookup table contains 5 million users and 5 million
items per each lookup table. As shown, larger embeddings rather than larger MLP dimension size cause a much more
dramatic increase in model size.
}
\vspace{-0.7em}
\label{fig:embedding_mem_usage}
\end{figure}

	 \begin{enumerate}

	 \item As embedding lookup tables are stored in the low-bandwidth CPU
	 memory, reading out embeddings (i.e., the embedding \emph{gather} operation) adds
	 significant latency than an unbuildable, oracular GPU-only version (\gpuonly) which assumes
	 infinite GPU memory capacity. Under \gpuonly, the entire embeddings can be stored locally in the high-bandwidth
	 GPU memory so gathering embeddings can be done much faster than when using CPU memory.	This
	 is because gathering embeddings is a memory bandwidth-limited
	 operation.

	 \item \cpuonly versions can sidestep the added latency that hybrid \cpugpu versions experience
	 during the PCIe communication process of transferring embeddings, but the (relatively) low computation throughput of
	 CPUs can significantly lengthen the DNN computation step.

	 \item Hybrid \cpugpu versions on the other hand can reduce the
	 DNN computation latency, but comes at the cost of additional CPU$\rightarrow$GPU
	 communication latency when copying the embeddings over PCIe using \texttt{cudaMemcpy}.

	 \end{enumerate}

 \subsection{Our Goal: A Scalable Memory System}
\label{sect:motivation_sec3}

Overall, the key challenge rises from the fact that the source operands of key
tensor manipulations (i.e., the embeddings subject for tensor concatenation or
		tensor reduction) are initially located inside the embedding lookup tables,
			 all of which are stored inside the capacity-optimized but bandwidth-limited
			 CPU memory. Consequently, prior solutions suffer from
			 \emph{low-bandwidth embedding gather operations} over CPU memory which adds significant latency.
			 Furthermore,
			  \cpuonly and \cpugpu versions need to trade-off the \emph{computational
				 bottleneck} of low-throughput CPUs over the \emph{communication
				 bottleneck} of PCIe, which adds additional latency overheads (\fig{fig:upcoming_dnns}).
			 What is more troubling is the
			 fact future projections of DL applications utilizing embedding layers
			 assume even larger number of embedding lookups and larger embeddings
			 themselves~\cite{park:2018:fb,hestness:2019:ppopp}, with complex tensor
			 manipulations to combine embedding features to improve
			 algorithmic performance.
			 			 Overall, both current
			 and future memory requirements of embedding layers point to an urgent
			 need for a system-level solution that provides \emph{scalable} memory
			 capacity and bandwidth expansion. In the next section, we detail our proposed
			 solution that addresses both the computational bottleneck of low-throughput CPUs
			 and the communication bottleneck of low-bandwidth CPU-GPU data transfers.

\section{TensorDIMM: An NMP DIMM Design for Embeddings \& Tensor Ops}
\label{sect:proposed}

\subsection{Proposed Approach}
\label{sect:opportunity_approach}

Our proposal is based on the following {\bf key observations} that open
up opportunities to tackle the system-level bottlenecks in
deploying memory-limited recommender systems.

\begin{figure}[t!] \centering
\hspace{0.1em}
\includegraphics[width=0.45\textwidth]{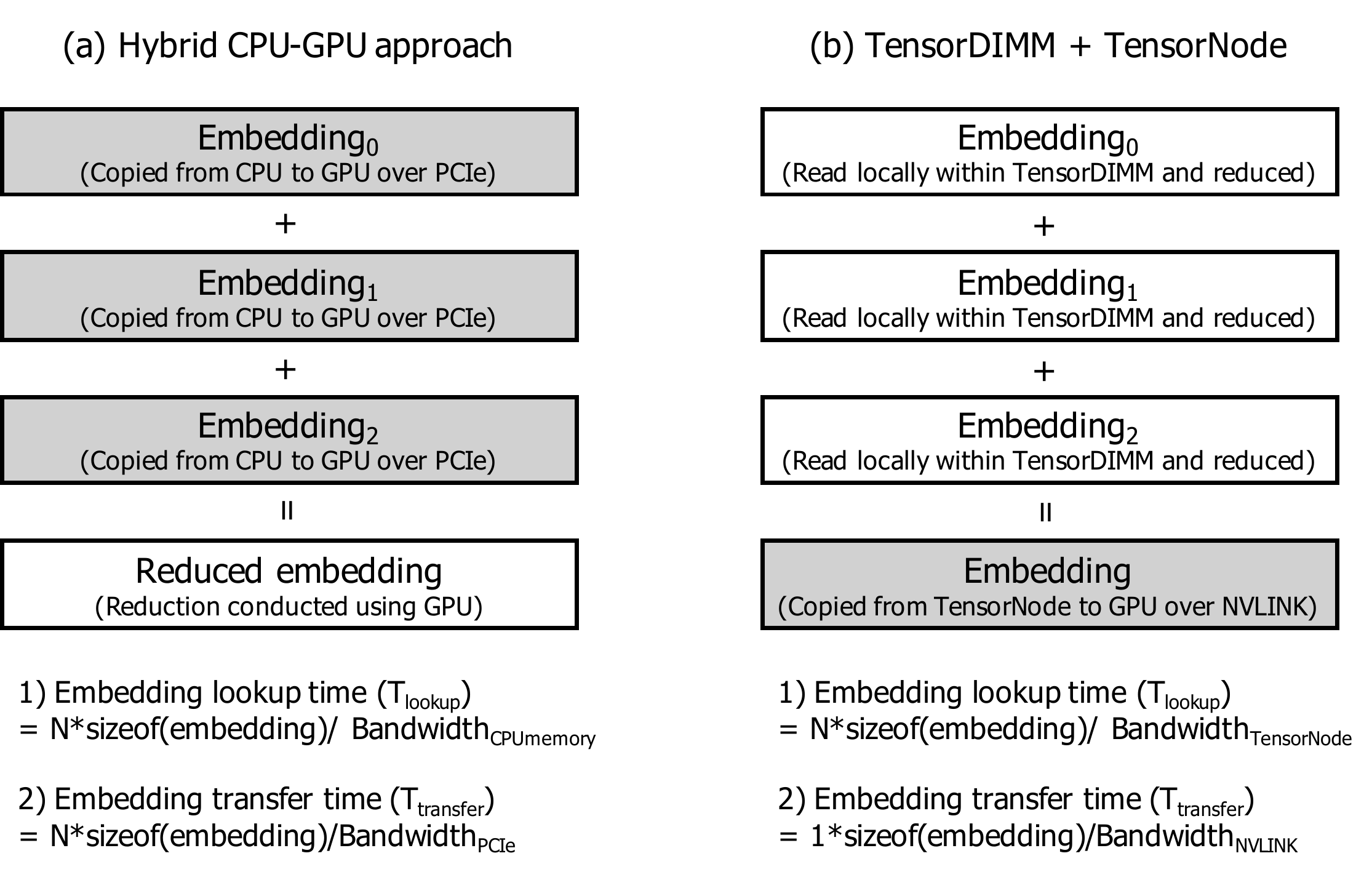}
\caption{
Our proposed approach: compared to the (a) hybrid \cpugpu version, (b) our solution 
	conducts embedding gathers and tensor-wide reductions locally ``near-memory'' first, and then
	transfers the \emph{reduced} tensor over the high-bandwidth NVLINK. Consequently, our proposal significantly
	reduces both 	the latency of gathering embeddings from the lookup table (\texttt{T$_{lookup}$})
	and the time taken to transfer the embeddings from the capacity-optimized CPU/\tnode
	memory to the bandwidth-optimized GPU memory (\texttt{T$_{transfer}$}). The \cpuonly version
	does not experience \texttt{T$_{transfer}$} but suffers from significantly longer DNN execution time
	than \cpugpu and our proposed design. 
}
\vspace{-0.5em}
\label{fig:proposed_approach}
\end{figure}

\begin{enumerate}

\item An important tensor operation used in combining embedding features
is the element-wise operation, which is equivalent to a tensor-wide
\emph{reduction} among \texttt{N} embeddings.  Rather than having the
\texttt{N} embeddings individually be gathered and copied over to the GPU memory for the
reduction operation to be initiated using the GPU, we can conduct the \texttt{N}
embedding gathers and reductions all ``near-memory'' and copy a single, \emph{reduced} tensor to
		GPU memory.  Such near-memory processing (NMP) approach reduces not only the latency to gather the embeddings but also the data transfer size
by a factor of \texttt{N} and alleviates the communication bottleneck of
CPU$\rightarrow$GPU \texttt{cudaMemcpy} (\fig{fig:proposed_approach}).

\item DL system vendors are deploying GPU/NPU-centric
interconnection fabrics~\cite{dgx_2,zion} that are decoupled from the legacy host-device PCIe.
Such technology allows vendors to employ custom high-bandwidth links
(e.g., NVLINK, providing $9\times$ higher bandwidth than PCIe) for fast
inter-GPU/NPU communication. But more importantly, it is possible to tightly
integrate ``non''-accelerator components (e.g., a disaggregated memory
		pool~\cite{disagg_mem_1,disagg_mem_2,mcdla}) as separate interconnect endpoints (or \emph{nodes}),
					allowing accelerators to read/write from/to these non-compute nodes using
					high-bandwidth links.

\end{enumerate}

	\begin{figure*}[t!] \centering
	\vspace{0.5em}
	\includegraphics[width=0.99\textwidth]{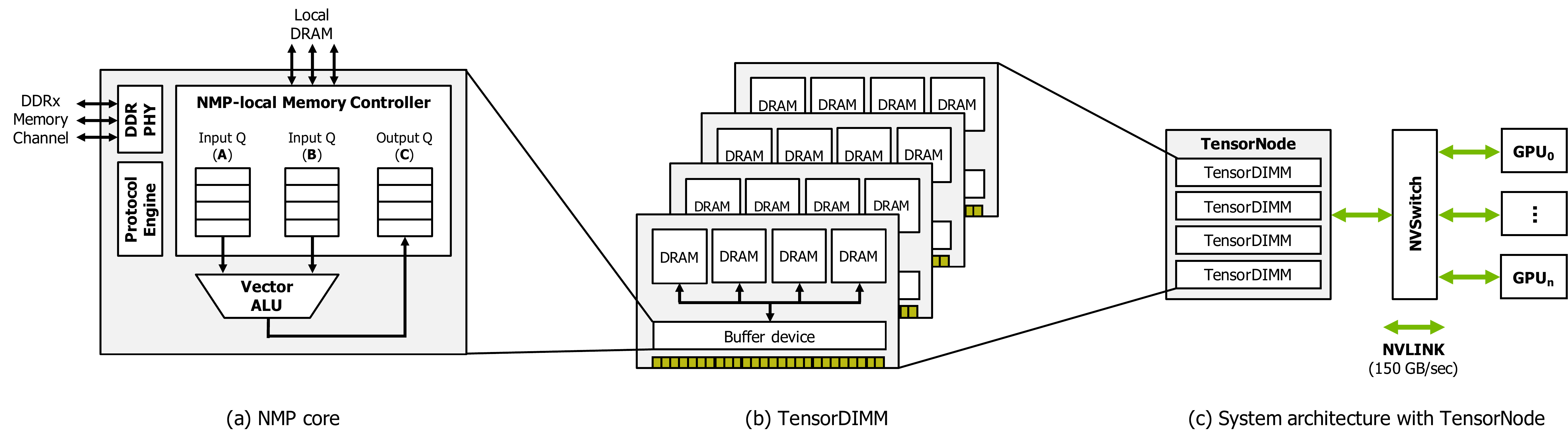}
\caption{
High-level overview of our proposed system. (a) An NMP core is implemented inside the buffer device
	of each (b) \tdimm, multiples of which are employed as a (c) disaggregated memory pool, called \tnode. The \tnode
	is integrated inside the high-bandwidth GPU-side interconnect. The combination of NVLINK and NVSwitch enables
	GPUs to read/write data to/from \tnode at a communication bandwidth up to $9\times$ higher
	than PCIe.
}
\label{fig:our_solution}
\end{figure*}

Based on these observations, we first propose \tdimm which is a custom DIMM
design including an NMP core tailored for tensor ``gather'' and
tensor ``reduction'' operations.  We then propose a disaggregated memory system called
\tnode which is fully populated with our \tdimms. Our last proposition is
a software architecture that effectively parallelizes the tensor gather/reduction
operation across the \tdimms, providing scalable memory bandwidth and capacity expansion.
The three most significant performance limiters
in conventional \cpuonly or hybrid \cpugpu recommender system are 1) the embedding gather
operation over the low-bandwidth CPU memory,  2) the compute-limited DNN execution using CPUs (for \cpuonly approaches)
	and 3) the CPU$\rightarrow$GPU
embedding copy operation over the PCIe bus (for hybrid \cpugpu), all of which adds severe latency
overheads (\sect{sect:recsys_bottleneck}).  Our vertically integrated solution
fundamentally addresses these problems thanks to the following three {\bf key innovations}
provided with our approach (\fig{fig:proposed_approach}).  First, 	by storing the entire embedding lookup
table inside the \tdimms, the embedding gather operation can be conducted
using the ample memory bandwidth available across the \tdimms ($4\times$ higher than CPU) rather than the low-bandwidth
CPU memory. Second, the \tdimm NMP cores conduct the \texttt{N} tensor reduction
operation \emph{before} sending them to the GPU, reducing the
\texttt{TensorNode}$\rightarrow$GPU communication latency by a factor of
\texttt{N}, effectively overcoming the communication bottleneck. Furthermore, the high-bandwidth
\texttt{TensorNode}$\leftrightarrow$GPU links (i.e., NVLINK) enable further
latency reduction, proportional to the bandwidth difference between PCIe and
NVLINK (approximately $9\times$). Lastly, all DNN computations are conducted
using the GPU, overcoming the computation bottleneck of \cpuonly implementations.
	As detailed in \sect{sect:evaluation}, our proposal achieves an average
	$6.2$$-$$15.0\times$ and $8.9$$-$$17.6\times$ performance improvement over the \cpuonly and hybrid \cpugpu, respectively, reaching
	$84\%$ of an unbuildable, oracular \gpuonly implementation with
	infinite memory capacity.  We detail each components of our proposal below.

\subsection{TensorDIMM for Near-Memory Tensor Ops}
\label{sect:tensor_dimm}

Our \tdimm is architected under three key design objectives in mind. First,
		\tdimm should leverage commodity DRAM chips as-is while be capable of being
		utilized as a normal buffered DIMM device in the event that it is not used
		for DL acceleration. Second, tensor reduction operations should be
		conducted ``near-memory'' using a lightweight NMP core, incurring minimum
		power and area overheads to the buffered DIMM device. Third, in addition to
		memory capacity, the amount of memory bandwidth available to the NMP cores
		should also scale up proportional to the number of \tdimm modules employed
		in the system.

{\bf Architecture.} \fig{fig:our_solution}(a,b) shows the \tdimm architecture, which
consists of an NMP core and the associated DRAM chips. As depicted, \tdimm does
not require any changes to commodity DRAMs since all modifications are
limited to a buffer device within a DIMM (\sect{sect:buffered_dimm}).  The NMP
core includes a DDR interface, a vector ALU, and an NMP-local
memory controller which includes input/output SRAM queues to stage-in/out the
source/destination operands of tensor operations.  The DDR interface
is implemented with a conventional DDR PHY and a protocol engine.

{\bf TensorDIMM usages.} For \emph{non}-DL use-cases, the
processor's memory controller sends/receives DRAM C/A and DQ signals to/from
this DDR interface, which directly interacts with the DRAM chips.  This allows
\tdimm to be applicable as a normal buffered DIMM device and be utilized by
conventional processor architectures for servicing load/store transactions.  Upon
receiving a \tisa instruction for tensor gather/reduction operations however,
					the instruction is forwarded to the NMP-local memory controller,
					which translates the \tisa instruction into low-level DRAM C/A
commands to be sent to the DRAM chips. Specifically, the
\tisa instruction is decoded (detailed in \sect{sect:tensor_isa}) in order to
calculate the physical memory address of the target tensor's location, and the NMP-local memory
controller generates the necessary \texttt{RAS/CAS/activate/precharge/}etc DRAM commands to
read/write data from/to the \tdimm DRAM chips.
 The data read out of DRAM chips are
temporarily stored inside the input ($A$ and $B$) SRAM queues until the ALU
reads them out for tensor operations.
In terms of NMP compute, the minimum
data access granularity over the eight x8 DRAM chips is $64$ bytes for a
burst length of $8$, which amounts to sixteen $4$-byte scalar elements.
As detailed in \sect{sect:tensor_isa}, the tensor operations accelerated with
our NMP cores are element-wise arithmetic operations (e.g., add, subtract,
		average, $\ldots$) which exhibit data-level parallelism across the
sixteen scalar elements.  We therefore employ a $16$-wide vector ALU which
conducts the element-wise operation over the data read out of the ($A$ and $B$)
	SRAM queues.  The vector ALU checks for any newly submitted pair of data
	($64$ bytes each) and if so, pops out a pair for tensor operations, the
	result of which is stored into the output ($C$) SRAM queue. The NMP memory
	controller checks the output queue for newly inserted results and drain them
	back into DRAM, finalizing the tensor reduction process. For tensor
	gathers, the NMP core forwards the data read out of the input
	queues to the output queue to be committed back into DRAM.

{\bf Implementation and overhead.}	\tdimm leverages existing DRAM chips and the associated
DDR PHY interface as-is, so the additional components introduced with our
\tdimm design are the NMP-local memory controller and the $16$-wide vector ALU. In
terms of the memory controller, decoding the \tisa instruction into a series of
DRAM commands is implemented as a FSM control logic so the major area/power
overheads comes from the input/output SRAM queues.  These buffers must be large
enough to hold the bandwidth-delay product of the memory sourcing the data to
remove idle periods in the output tensor generation stream. A conservative
estimate of $20$ ns latency from the time the NMP-local memory
controller requests data to the time it arrives at the SRAM queue is used to
size the SRAM queue capacity.  Assuming the baseline PC4-25600 DIMM that provide $25.6$
GB/sec of memory bandwidth, the NMP core requires ($25.6$ GB/sec$\times$$20$
		ns) = $0.5$ KB of SRAM queue size ($1.5$ KB overall, for both input/output
			queues).  The $16$-wide vector ALU is clocked at $150$ MHz to
		provide enough computation throughput to seamlessly conduct the
		element-wise tensor operations over the data read out of the input ($A$ and $B$)
	queues.  Note that the size of an IBM Centaur buffer
	device~\cite{centaur} is approximately ($10$mm$\times$$10$mm) with a TDP of $20$ W.
	Compared to such design point, our NMP core adds negligible power and area
	overheads, which we quantitatively evaluate in \sect{sect:impl_overhead}.

{\bf Memory bandwidth scaling.} An important design objective of \tdimm is to
provide scalable memory bandwidth expansion for NMP tensor operations.  A key
challenge with conventional memory systems is that the maximum bandwidth per
each memory channel is fixed (i.e., signaling bandwidth per each pin $\times$
		number of data pins per channel), regardless of the number of DIMMs (or
			ranks/DIMM) per channel.  For instance, the maximum CPU memory bandwidth
		available under the baseline CPU system (i.e., NVIDIA DGX~\cite{dgx_1v})
	can never exceed ($25.6$ GB/sec$\times$8)$=$$204.8$ GB/sec across the eight
	memory channels ($4$ channels per each socket), irrespective of the number of
	DIMMs actually utilized (i.e., $8$ DIMMs vs. $32$ DIMMs).  This is because
	the physical memory channel bandwidth is time-multiplexed across multiple
	DIMMs/ranks.   As detailed in the next subsection, our \tdimm is utilized as a
	basic building block in constructing a disaggregated memory system (i.e.,
			\tnode).  The key innovation of our proposal is that, combined with our
	\tisa address mapping function (\sect{sect:tensor_isa}), the amount of
	aggregate memory bandwidth provided to all the NMP cores within \tnode
	increases proportional to the number of \tdimm employed: an aggregate of
	$819.2$ GB/sec memory bandwidth assuming $32$ \tdimms, a $4\times$ increase
	over the baseline CPU memory system. Such memory bandwidth scaling is
	possible because the NMP cores access its \tdimm-internal DRAM chips
	``locally'' within its DIMM, not having to share its local memory bandwidth
	with other \tdimms. Naturally, the more \tdimms employed inside the memory
	pool, the larger the aggregate memory bandwidth becomes available to the NMP
	cores conducting embedding gathers and reductions.

\subsection{System Architecture}
\label{sect:tensor_node}

\fig{fig:our_solution} provides a high-level overview of our proposed system
architecture.  We propose to construct a disaggregated memory pool within the
high-bandwidth GPU system interconnect. Our design is referred to as \tnode because it
functions as an interconnect endpoint, or \emph{node}. GPUs can read/write data
from/to \tnode over the NVLINK compliant PHY interface, either using
fine-grained CC-NUMA or coarse-grained data transfers using
P2P \texttt{cudaMemcpy}\footnote{CC-NUMA access
or P2P \texttt{cudaMemcpy} among NVLINK-compatible devices is already
available in commercial systems (e.g., Power9~\cite{ibm_power9}, GPUs within DGX-2~\cite{dgx_2}). Our \tnode
leverages such technology as-is to minimize design complexity.
}.
\fig{fig:our_solution}(c) depicts our \tnode design populated with multiple \tdimm
devices.  The key advantage of \tnode is threefold. First, \tnode provides a
platform for increasing memory capacity in a \emph{scalable} manner as the disaggregated
memory pool can independently be expanded  using density-optimized DDRx, irrespective of the
	GPU's local, bandwidth-optimized (but capacity-limited) 3D stacked memory.  As such,
	it is possible to store multiples embedding lookup tables entirely inside
	\tnode because the multitude of \tdimm devices (each equipped with
			density-optimized LR-DIMM~\cite{lrdimm}) enable scalable memory capacity increase.
	Second, the aggregate  memory bandwidth  available to the \tdimm
	NMP cores has been designed to scale up proportional to the number of DIMMs provisioned within the
	\tnode (\sect{sect:tensor_isa}). This allows our \tnode and \tdimm design to
	fulfill not only the current but also future memory (capacity and bandwidth) needs of recommender
	systems which combines \emph{multiple} embeddings, the size of which
	is expected to become even \emph{larger} moving forward (\sect{sect:recsys_bottleneck}). Recent projections from several hyperscalars~\cite{park:2018:fb,hestness:2019:ppopp}
	state that the memory capacity requirements of embedding layers will increase by hundreds of times larger
	than the already hundreds of GBs of memory footprint. \tnode is a scalable,
	future-proof system-level solution that addresses the memory bottlenecks of embedding layers.
	Third, the communication channels to/from the
	\tnode is implemented using high-bandwidth NVLINK PHYs so transferring
	embeddings between a GPU and a \tnode becomes much faster than when using
	PCIe (approximately $9\times$).

\subsection{Software Architecture}
\label{sect:tensor_isa}

We now discuss the software architecture of our proposal: the address mapping scheme for
embeddings, \tisa for
conducting near-memory operations, and the runtime system.

\begin{figure}[t!] \centering
\hspace{-1.5em}
\subfloat[]{
	\includegraphics[width=0.485\textwidth]{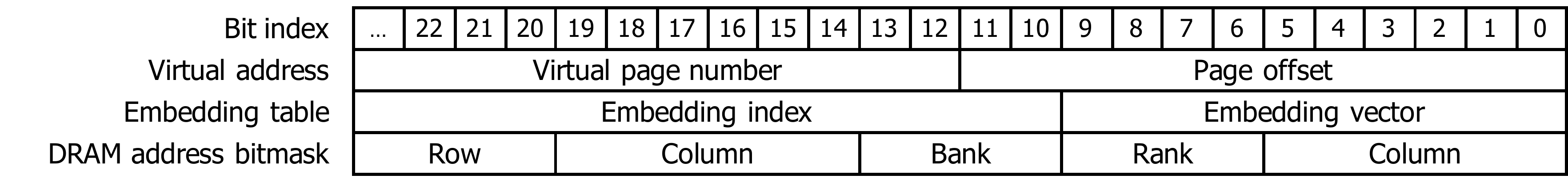}
}
\vspace{0em}
\subfloat[]{
	\includegraphics[width=0.36\textwidth]{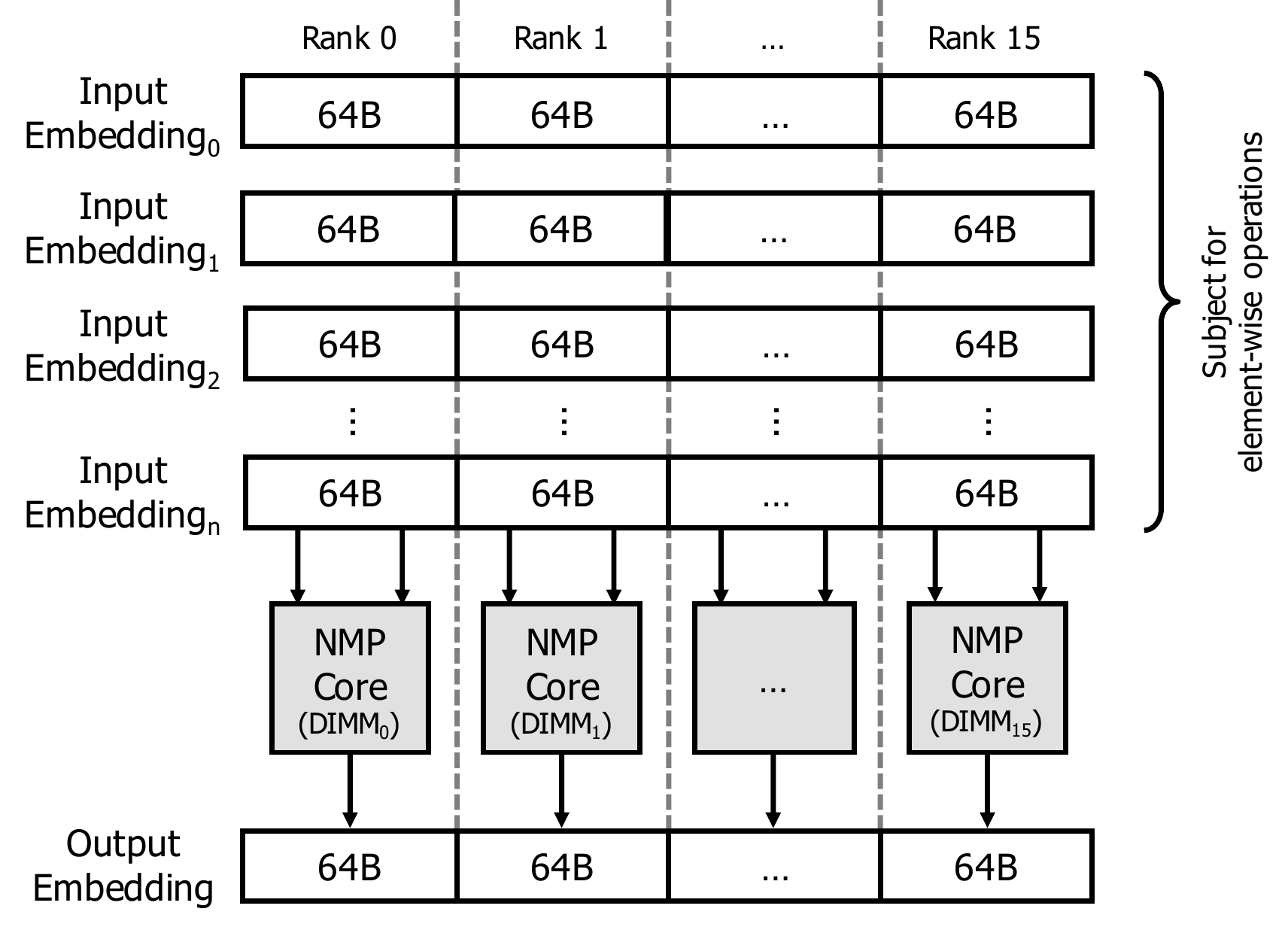}
}
\vspace{0em}
\caption{
(a) Proposed DRAM address mapping scheme for embeddings, assuming
each embedding is $1$ KB ($256$-dimension) and is split evenly across $16$ ranks (or DIMMs) (b)
Example showing how rank-level parallelism is utilized to interleave and map each
embedding across the $16$ \tdimms. Such rank-level parallelism centric address mapping scheme
enable the memory bandwidth available to the NMP cores to increase proportional to the number
of \tdimms employed.
}
\label{fig:tensor_map}
\end{figure}

{\bf Address mapping architecture.} One of the key objectives of our address
mapping function is to provide scalable performance improvement whenever
additional \tdimms (i.e., number of DIMMs$=$NMP cores) are added to our \tnode.
To achieve this goal, it is important that all NMP cores within the \tnode
concurrently work on a distinct subset of the embedding vectors for
(gather/reduce) tensor operations.  \fig{fig:tensor_map}(a) illustrates our
address mapping scheme which utilizes rank-level parallelism to maximally
utilize both the \tdimm's NMP computation throughput and memory bandwidth.
Because each \tdimm has its own NMP core, maximally utilizing aggregate NMP
compute throughput requires all \tdimms to work in parallel. Our address
mapping function accomplishes this by having consecutive $64$ bytes\footnote{We
	assume \tdimm is built using a x64 DIMM. With a burst length of $8$, the
		minimum data access granularity becomes $64$ bytes.} within each embedding
		vector be interleaved across different ranks, allowing each \tdimm to
		independently work on its own slice of tensor operation concurrently
		(\fig{fig:tensor_map}(b)).  As our target algorithm contains abundant
		data-level parallelism, our address mapping technique effectively
		partitions and load-balances the tensor operation across all \tdimms
		within the \tnode. In \sect{sect:eval_mem_bw}, we quantitatively evaluate
		the efficacy of our address mapping function in terms of maximum DRAM
		bandwidth utilization.

							 It is worth pointing out that our address mapping function is
							 designed to address not just the current but also future
							 projections on how DL practitioners seek to use embeddings.
							As discussed in \sect{sect:recsys_bottleneck}, given the
							 capability, DL practitioners are willing to adopt \emph{larger}
							 embedding vector dimensions to improve the algorithmic performance. This
							 translates into a larger number of bits to reference any given
							 embedding vector (e.g., enlarging embedding size from $1$ KB to
									 $4$ KB increases the embedding's bit-width from $10$ bits to
									 $12$ bits in \fig{fig:tensor_map}(a)) leading to both larger
							 memory footprint to store the embeddings and higher
							 computation and memory bandwidth demands to conduct tensor
							 reductions.  Our system-level solution	has been
							 designed from the ground-up to effectively handle such user
							 needs: that is, system architects can provision more \tdimm
							 ranks within \tnode and increase its memory capacity
							 proportional to the increase in embedding size, which is
							 naturally accompanied by an increase in NMP compute throughput
							 \emph{and} memory bandwidth simultaneously.

\begin{figure}[t!] \centering
	\includegraphics[width=0.48\textwidth]{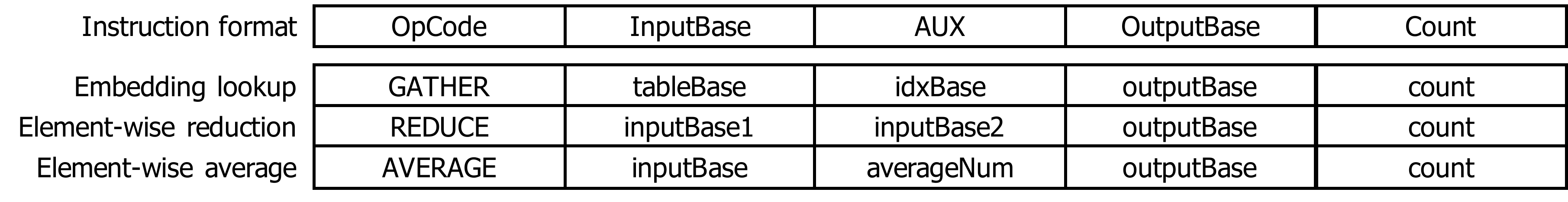}
\caption{
Instruction formats for \lookup, \reduce, and \average.
}
\vspace{-1em}
\label{fig:tensor_isa}
\end{figure}

\begin{figure}[t!] \centering
\subfloat[]
{
	\includegraphics[width=0.50\textwidth]{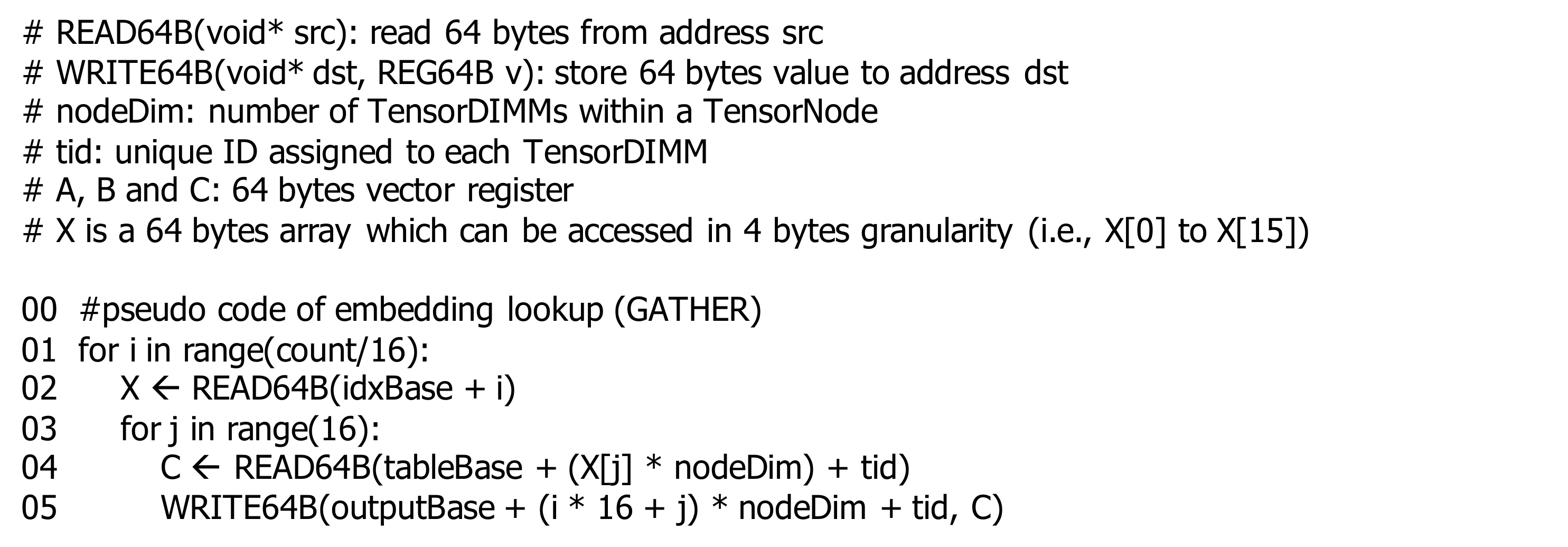}
}
\vspace{0em}
\subfloat[]
{
	\includegraphics[width=0.50\textwidth]{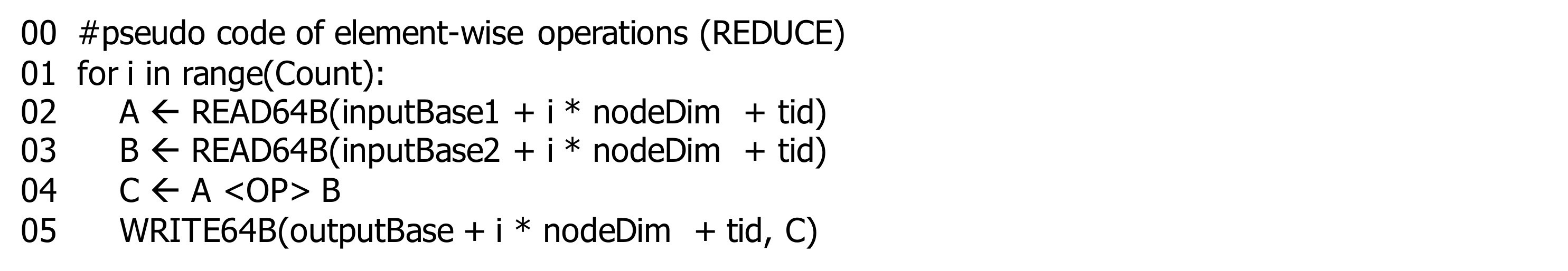}
}
\vspace{0em}
\subfloat[]
{
	\includegraphics[width=0.50\textwidth]{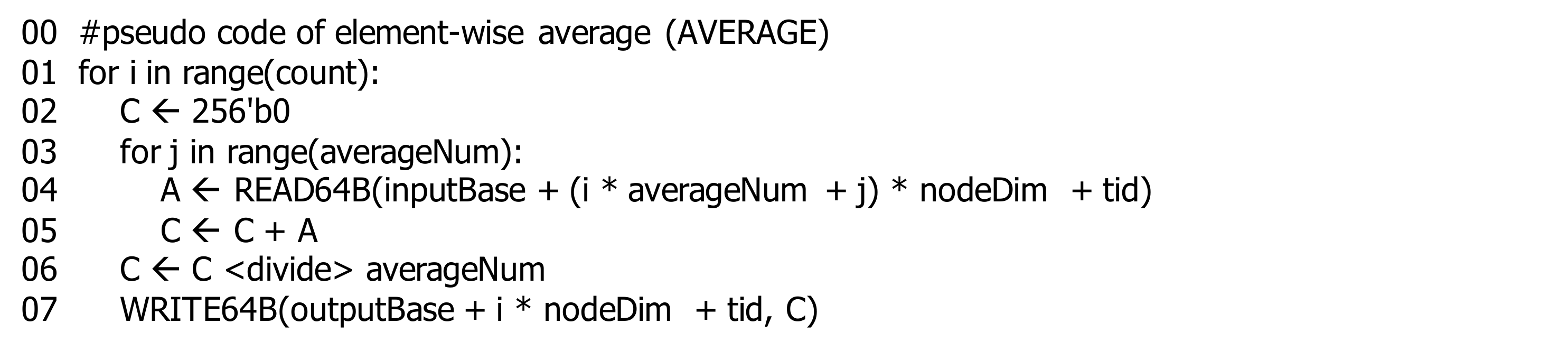}
}
\vspace{0em}
\caption{
Pseudo code explaining the functional behavior of (a) \lookup, (b) \reduce, and (c) \average.
}
\label{fig:tisa_code}
\end{figure}

{\bf TensorISA.} The near-memory tensor operations are initiated using our custom
ISA extension called \tisa. There are three key \tisa primitives supported in
\tdimm: the \lookup instruction for embedding lookups and the \reduce and
\average instructions for element-wise operations.  \fig{fig:tensor_isa} and
\fig{fig:tisa_code} summarize the instruction formats of these three
instructions and pseudo codes describing each of its functional behavior.
Consider the example in \fig{fig:ncf} which assumes an embedding layer that
uses two embedding lookup tables with a batch size $4$ to compose two tensors,
		 followed by an element-wise operation among these two for reduction. As
		 discussed in \sect{sect:embedding}, an embedding layer starts with an
		 embedding lookup phase that \emph{gathers} multiple embeddings up to the
		 batch size from the embedding lookup table, followed by various tensor
		 manipulations.  Under our proposed system, the GPU executes this embedding
		 layer as follows.  First, the GPU sends three instructions, two \lookup
		 instructions and one \reduce, to the \tnode.  The \tisa instruction is
		 broadcasted to all the \tdimms because each NMP core is responsible for
		 locally conducting its share of the embedding lookups as well as its slice of the
		 tensor operation (\fig{fig:tisa_code}(a,b)).  For instance, assuming the address
		 mapping function in \fig{fig:tensor_map}(a) and a \tnode configuration
		 with $16$ \tdimms, a single \lookup instruction will have each \tdimm
		 gather ($4$$\times$$64$ bytes) = $256$ bytes of data under a
		 contiguous physical address space, the process of which is orchestrated by
		 the \tdimm NMP-local memory controller using DRAM read/write transactions
		 (\sect{sect:tensor_dimm}).  The \lookup process is undertaken twice to
		 prepare for the two tensor slices per each \tdimm rank. Once the two
		 tensors are gathered, a \reduce instruction is executed by the NMP cores (\sect{sect:tensor_dimm}).

{\bf Runtime system.} DL applications are typically encapsulated as a direct
acyclic graph (DAG) data structure in major DL
frameworks~\cite{tensorflow,torch,mxnet}. Each node within the DAG represents a
DNN layer and the DL framework compiles down the DAG into sequence of host-side
CUDA kernel launches that the GPU executes one layer at a time.  The focus of
this paper is on recommender systems which utilizes embedding lookups and
various tensor manipulations. Under our proposed system, embedding layers are
still executed using normal CUDA kernel launches but the kernel itself is wrapped around specific
information for our \tdimm runtime system to utilize for near-memory tensor operations.
Specifically, as part of the embedding layer's CUDA kernel context,
	information such as the number of table lookups, the embedding dimension size,
	tensor reduction type, the	input batch size, and etc are encoded per \tisa instruction's format (\fig{fig:tensor_isa})
	and is sent to the GPU as part of the CUDA kernel launch. When the GPU runtime receives
	these instructions, they are forwarded to the \tnode for near-memory processing as discussed
	in \sect{sect:tensor_dimm}.

	As \tnode is a remote, disaggregated memory pool from the GPU's perspective,
	the runtime system  should be able to (de)allocate memory inside this remote
	memory pool. Our proposal builds upon our prior work~\cite{mcdla} which proposes several CUDA runtime API
	extensions for remote memory (de)allocation under a GPU-side disaggregated
	memory system. We refer the interested readers to
	\cite{mcdla,buddy_compression} for further details on the runtime CUDA APIs
	required for memory (de)allocations within a disaggregated memory system.

\section{Evaluation Methodology}
\label{sect:methodology}

Architectural exploration of \tdimm and its system-level implication within
\tnode using a cycle-level simulator is challenging for several reasons. First,
	running a single batch inference for DL applications can take up to several
	milliseconds even on high-end GPUs, so running cycle-level simulation on
	several tens to hundreds of batches of inference leads to intractable amount
	of simulation time. Second, our proposal covers multiple levels in the hardware/software
	stack, so a cycle-level hardware performance model of \tdimm and
	\tnode alone will not properly reflect the complex interaction of
	(micro)architecture, runtime system, and the system software, potentially
	resulting in misleading conclusions.  Interestingly, we note that the key DL
	operations utilized in embedding layers that are of interest to this study are
	completely \emph{memory bandwidth limited}. This allows us to utilize the existing
	DL hardware/software systems to ``emulate'' the behavior of \tdimm and
	\tnode on top of state-of-the-art real DL systems. Recall that the embedding
	lookups (\lookup) and the tensor reduction operations (\reduce/\average) have
	extremely low compute-to-memory ratio (i.e., all three operations
			are effectively \emph{streaming} applications), rendering the execution of these three
	operations to be bottlenecked by the available memory bandwidth.
	Consequently, the effectiveness of our proposal primarily lies in 1) the
	effectiveness of our address mapping scheme on maximally utilizing the aggregate
	memory bandwidth for the tensor operations across
	all the \tdimms, and 2) the impact of ``PCIe vs. NVLINK''  on the
	communication latency of copying embeddings across the capacity-optimized
	``CPU vs. \tnode'' memory and bandwidth-optimized GPU memory (\fig{fig:proposed_approach}). We introduce
	our novel, hybrid evaluation methodology that utilizes both cycle-level simulation and
	a proof-of-concept prototype developed on real DL systems
	to quantitatively demonstrate the benefits of our proposal.

\begin{table}[t!]
\centering
\caption{Baseline \tnode configuration.}
\footnotesize
  \begin{tabular}{|c|c|}
		\hline
    DRAM specification     			&  DDR4 (PC4-25600)  \\
		\hline
    Number of \tdimms     			&   $32$ \\
		\hline
    Memory bandwidth per \tdimm     			& $25.6$ GB/sec   \\
		\hline
    Memory bandwidth across \tnode     			& $819.2$ GB/sec   \\
    \hline
  \end{tabular}
\vspace{-0.5em}
  \label{tab:dram_config}
\end{table}

{\bf Cycle-level simulation.} As the performance of \tdimm and \tnode is
bounded by how well they utilize the DRAM bandwidth, an evaluation of
our proposal on memory bandwidth utilization is in need.
We develop a memory
tracing function that hooks into the DL frameworks~\cite{tensorflow,torch} to
generate the necessary read/write memory transactions in executing
\lookup, \reduce, and \average operations for embedding lookups and tensor
operations. The traces are fed into Ramulator~\cite{ramulator}, a
cycle-accurate DRAM simulator, which is configured to model 1) the baseline
CPU-GPU system configuration with eight CPU-side memory channels, and
2) our proposed
address mapping function (\sect{sect:tensor_isa}) and \tnode configuration (\tab{tab:dram_config}),
which we utilize to measure the effective
memory bandwidth utilization when executing the three tensor operations
under baseline and \tnode.

\begin{figure}[t!] \centering
\hspace{4em}	\includegraphics[width=0.45\textwidth]{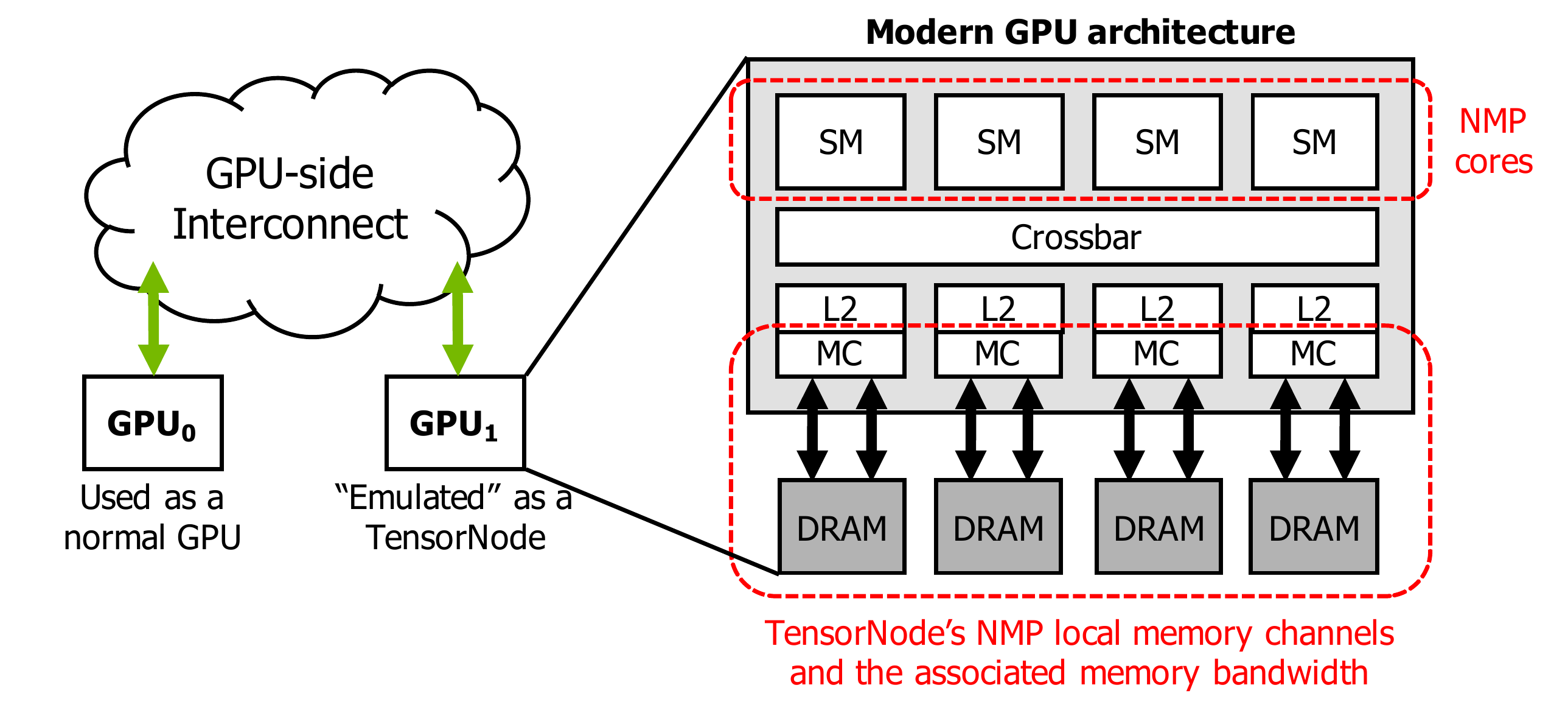}
\caption{
Emulation of \tdimm and \tnode using a real GPU: the GPU cores (aka SMs~\cite{cuda}) and the GPU-local
	memory channels (and bandwidth) corresponds to the NMP cores within \tdimms and the aggregate
	memory bandwidth available across the \tnode, respectively.
}
\vspace{-0.5em}
\label{fig:prototype}
\end{figure}

{\bf ``Proof-of-concept'' prototype.} We emulate the behavior
of our proposed system using the state-of-the-art NVIDIA
DGX~\cite{dgx_1v} machine, which includes eight V100 GPUs~\cite{volta_v100}.
Each V100 GPU contains $900$ GB/sec of local memory bandwidth with six
NVLINKs for communicating with other GPUs up to $150$
GB/sec~\cite{nvlink}. To emulate the system-level effects of the high-bandwidth
NVLINK communication channels between \texttt{TensorNode}$\leftrightarrow$GPU,
			 we use a pair of GPUs in DGX and treat one of them as our proposed
			 \tnode while the other acts as a normal GPU (\fig{fig:prototype}).  Because the
			 tensor operations accelerated using our \tdimms are memory bandwidth-limited, streaming workloads, the performance difference between a
			 (hypothetical) real \tnode populated with \texttt{N} \tdimms and a
			 single V100 that we emulate as our \tnode will be small, provided the
			 V100 local memory bandwidth matches that of our assumed \tnode
			 configuration. As such, when evaluating the effective memory bandwidth
			 utilization of \tnode, we configured the number of ranks (i.e., \tdimms)
	to be \texttt{N}=$32$ such that the aggregate memory bandwidth available
	within a single \tnode approximately matches that of a single V100 ($900$
			GB/sec vs.  $819.2$ GB/sec, \tab{tab:dram_config}).  After validating the
	effectiveness of \tnode in utilizing memory bandwidth, commensurate to that
	of V100, we implement a software prototype of an end-to-end recommender
	system (configurable to the four DL applications discussed below) using
	Intel's Math Kernel Library (MKL version $2019.0.3$~\cite{mkl}), cuDNN
	(version $7$~\cite{cudnn}), cuBLAS~\cite{cublas}, and our in-house
	CUDA implementation of embedding layers (including \lookup/\reduce/\average)
	as well as other layers that do not come with MKL, cuDNN or cuBLAS. We
	cross-validated our in-house implementation of these memory bandwidth-limited
	layers by comparing the measured performance against the upper bound, ideal
	performance, which exhibited little variation.  Under our CUDA implementation
	of tensor reduction operations, the GPU cores (called SMs in
			CUDA~\cite{cuda}) effectively function as the NMP cores within \tnode
	because the CUDA kernels of \reduce/\average stage in/out the tensors between
	(SM$\leftrightarrow$GPU local memory) in a streaming fashion as done in
	\tdimm.  The \texttt{TensorNode}$\leftrightarrow$GPU communication is
	orchestrated using P2P \texttt{cudaMemcpy} over NVLINK when evaluating our
	proposal. When running the sensitivity of \tdimm on
	(\texttt{TensorNode}$\leftrightarrow$GPU) communication bandwidth, we
	artificially increase (decrease) the data transfer size to emulate the
	behavior of smaller (larger) communication bandwidth and study its
	implication on system performance (\sect{sect:sensitivity_comm_bw}).

{\bf Benchmarks.} We choose four neural network based recommender system
applications using embeddings to evaluate our proposal: (1) the neural
	collaborative filtering (\texttt{NCF})~\cite{ncf} based recommender
		system available in MLPerf~\cite{mlperf}, (2) the YouTube recommendation
		system~\cite{youtube_recsys} (\texttt{YouTube}), and (3) the Fox movie
		recommendation system~\cite{fox} (\texttt{Fox}),
		and the Facebook recommendation system~\cite{facebook_dlrm} (\texttt{Facebook}\footnote{Facebook's
		Deep Learning Recommendation Model~\cite{facebook_dlrm} was released \emph{after} this work was submitted 
		for peer-review at MICRO-52. As such, we implement and evaluate this model using our proof-of-concept emulation framework, rather than
		using Facebook's open-sourced PyTorch~\cite{torch} version, for consistency with the other three models.}).
		To choose a realistic
		inference batch size representative of real-world inference scenarios, we refer to
		the recent study from Facebook~\cite{park:2018:fb} which states that
		datacenter recommender systems are commonly deployed with a batch size of
		$1$$-$$100$. Based on this prior work, we use a batch size of $64$ as our
		default configuration but sweep from batch $1$ to $128$ when running sensitivity studies.
		All four workloads are configured with a default embedding vector
		dimension size of $512$ with other key application configurations
		summarized in \tab{tab:benchmarks}. We specifically note when deviating from these
		default configurations when running sensitivity studies in \sect{sect:sensitivity_large_embedding}.

{\bf Area/power.} The implementation overheads of \tdimm
are measured with synthesized implementations using
Verilog HDL, targeting a Xilinx Virtex UltraScale+ VCU1525 acceleration dev board.
The system-level power overheads of \tnode are evaluated using Micron's DDR4 power calculator~\cite{micron_dram_power_estimator}.
We detail these results in \sect{sect:impl_overhead}.

\begin{table}[t!]
  \centering
  \caption{Evaluated benchmarks and default configuration.}
\footnotesize
  \begin{tabular}{|c|c|c|c|}
    \hline
    \textbf{Network} & \textbf{Lookup tables} & \textbf{Max reduction} & \textbf{FC/MLP layers}\\
    \hline
    \hline
    \texttt{NCF}						&		$4$ & $2$	& 4 \\
    \hline
		\texttt{YouTube}  			&		$2$ & $50$ & 4\\
    \hline
    \texttt{Fox}   					&   $2$ & $50$	& 1\\
		\hline
    \texttt{Facebook}   		&   $8$ & $25$	& 6\\
    \hline
  \end{tabular}
\vspace{-0.5em}
  \label{tab:benchmarks}
\end{table}

\section {Evaluation} \label{sect:evaluation}

	We explore five design points of recommender systems: the 1) CPU-only
	version (\cpuonly), 2) hybrid CPU-GPU  version (\cpugpu), 3) \tnode style
	pooled memory interfaced inside the high-bandwidth GPU interconnect
	but utilizes regular capacity-optimized DIMMs, rather than the NMP-enabled
	\tdimms (\pmem), 4) our proposed \tnode with \tdimms (\tdm), and 5) an
	unbuildable, oracular GPU-only version (\gpuonly) which assumes that the
	entire embeddings can be stored inside GPU local memory, obviating the need
	for \texttt{cudaMemcpy}.

\begin{figure}[t!] \centering
\includegraphics[width=0.475\textwidth]{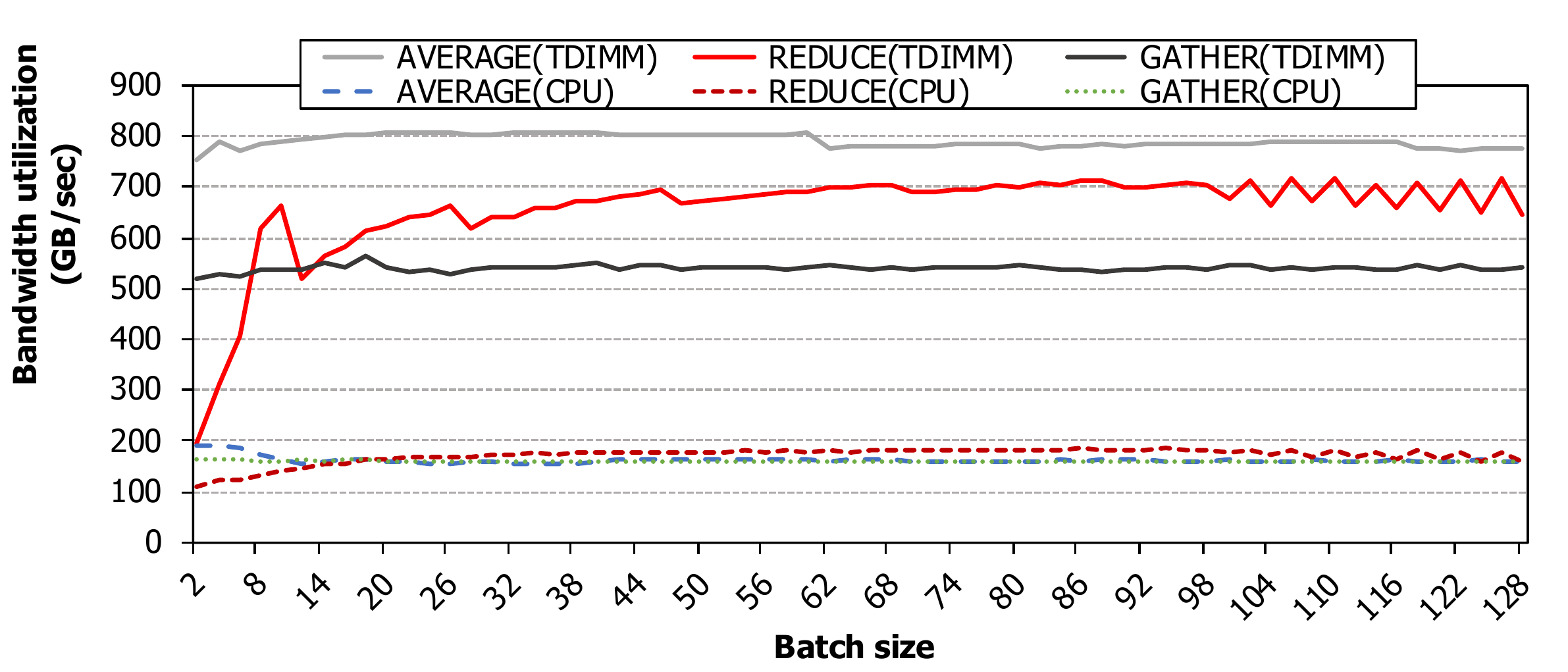}
\caption{
Memory bandwidth utilization for the three tensor operations. \tnode assumes
the default configuration in \tab{tab:dram_config} with 32 \tdimms. For \cpuonly and \cpugpu, the tensor operations
are conducted over a conventional CPU memory system, so a total
of 8 memory channels with 32 DIMMs (4 ranks per each memory channel) are assumed.
}
\label{fig:dram_bw_util}
\end{figure}

\begin{figure}[t!] \centering
\includegraphics[width=0.48\textwidth]{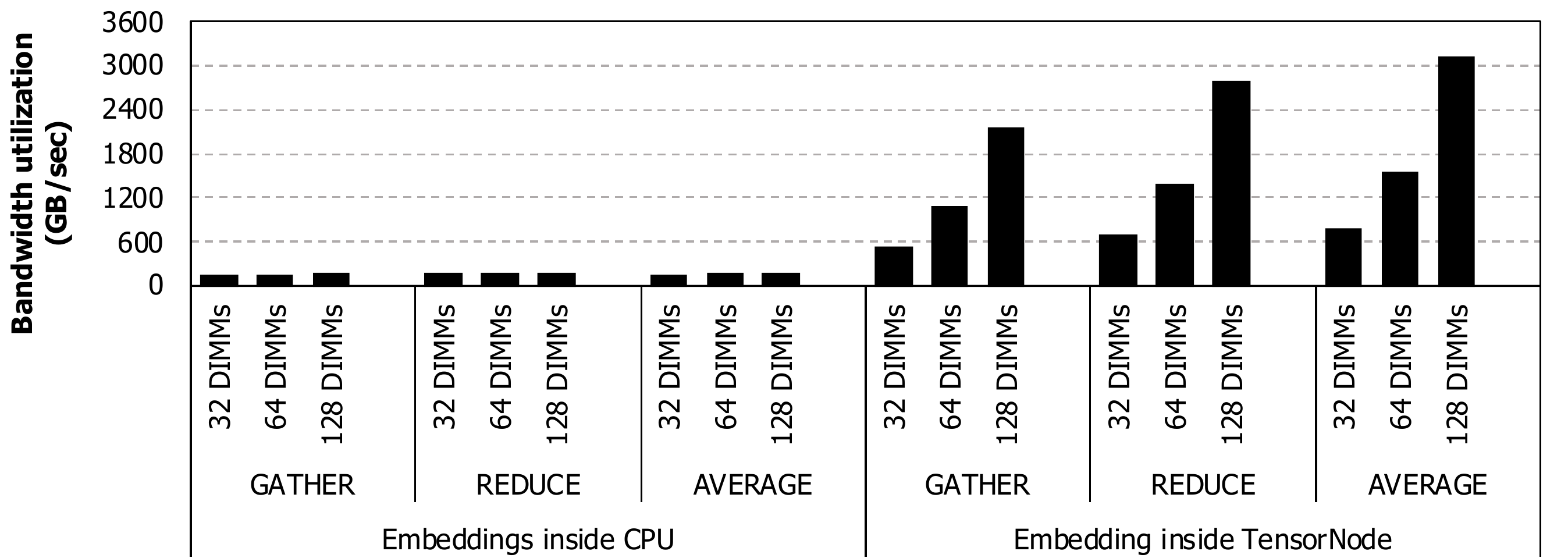}
\caption{
Memory throughput as a function of the number of DIMMs employed within CPU and \tnode. Evaluation
	assumes the embedding size is increased from the default value by 2$-$4$\times$, which
	proportionally increases the embedding lookup table size, requiring
a proportional
	increase in memory capacity (i.e., more DIMMs) to store these lookup tables.
}
\label{fig:dram_bw_scaling}
\end{figure}

\subsection{Memory Bandwidth Utilization}
\label{sect:eval_mem_bw}

To validate the effectiveness of \tnode in amplifying effective memory
bandwidth, we measure the aggregate memory
throughput achieved with \tnode and the baseline CPU-based systems, both of which
utilizes a total of $32$ DIMMs (\fig{fig:dram_bw_util}).  \tnode  significantly outperforms
the baseline CPU system with an average $4\times$ increase in memory bandwidth
utilization (max $808$ vs. $192$ GB/sec).  As discussed in \sect{sect:tensor_dimm},
						conventional memory systems time-multiplex the memory
						channel across multiple DIMMs, so larger number of DIMMs
						only provide enlarged memory capacity not bandwidth. Our \tdimm is
						designed to ensure that the aggregate memory bandwidth scales
						proportional to the number of \tdimms, achieving
						significant memory bandwidth scaling. The benefits of our
						proposal is more pronounced for more future-looking scenarios with
						enlarged embedding dimension sizes. \fig{fig:dram_bw_scaling} shows
						the effective memory bandwidth for tensor operations when
						the embedding dimension size is increased by up to $4\times$,
						necessitating larger number of DIMMs to house the proportionally
						increased embedding lookup table. As depicted, the baseline CPU memory system's
					memory bandwidth saturates at around $200$ GB/sec because
						of the fundamental limits of conventional memory systems.
						\tnode on the other hand reaches up to $3.1$ TB/sec
						of bandwidth, more than $15\times$ increase than baseline.

\subsection{System-level Performance}
\label{sect:perf}

\begin{figure}[t!] \centering
\includegraphics[width=0.485\textwidth]{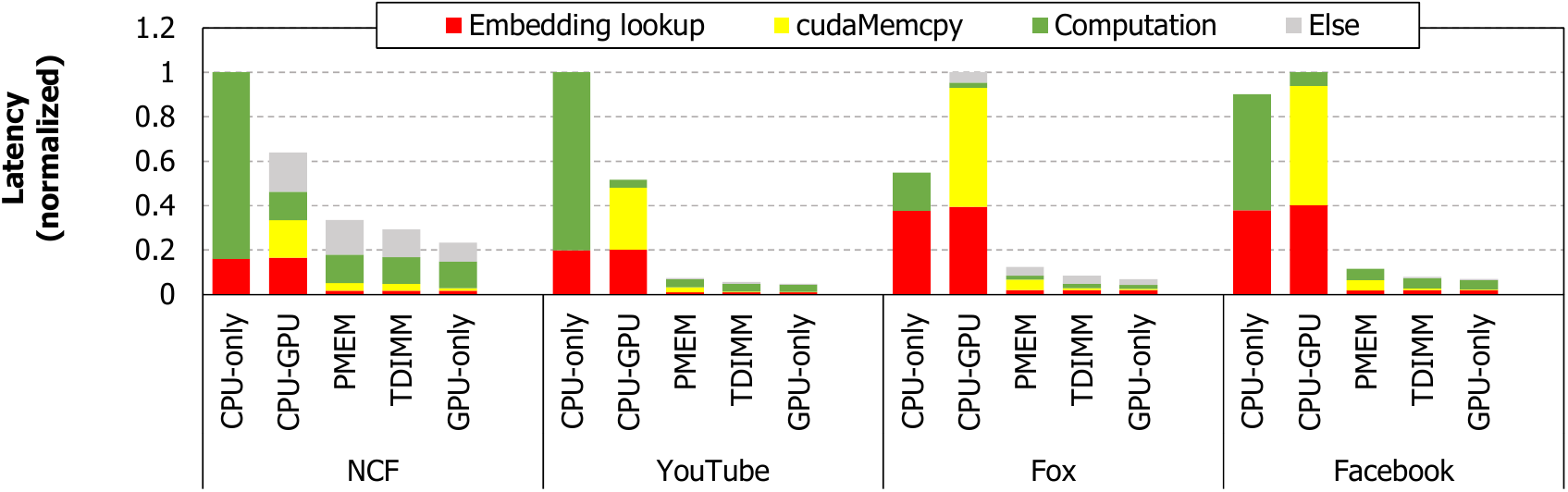}
\caption{
Breakdown of latencies during an inference with batch 64, normalized to
	the slowest design point (i.e., \cpuonly or \cpugpu).
}
\vspace{-0.5em}
\label{fig:latency_breakdown}
\end{figure}

\begin{figure*}[t!] \centering
\includegraphics[width=1.0\textwidth]{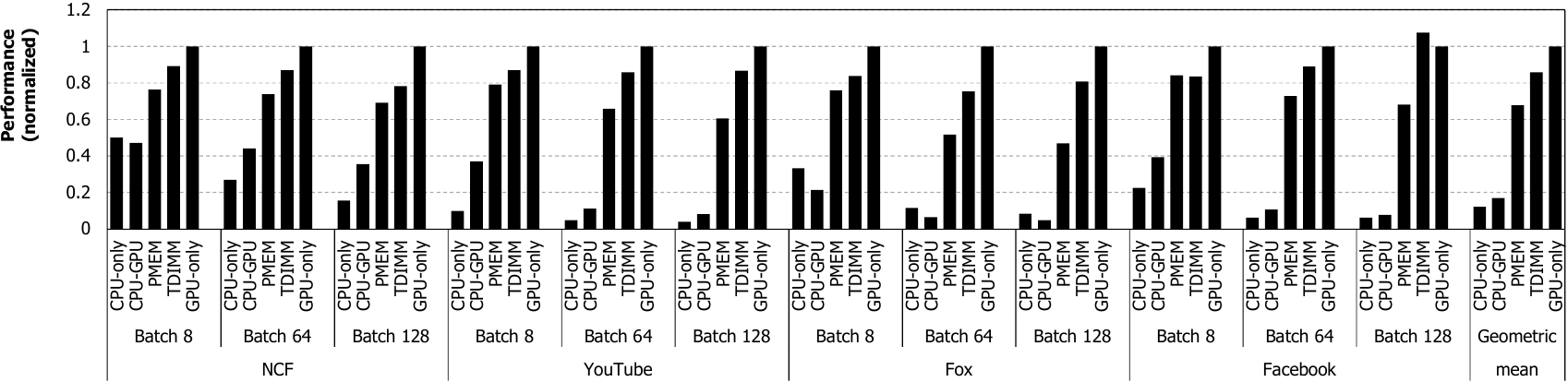}
\caption{
Performance of the five design points of recommender systems, normalized to the oracular GPU (\gpuonly).
}
\vspace{-0.5em}
\label{fig:perf_overall}
\end{figure*}

\tdimm significantly reduces the latency to execute memory-limited embedding layers,
	thanks to its high memory throughput and the communication bandwidth
amplification effects of both near-memory tensor operations \emph{and} the
high-bandwidth NVLINK. \fig{fig:latency_breakdown} shows a
latency breakdown of our studied workloads assuming a batch size of $64$.
With our proposed solution, all four applications enjoy significant reduction in both
embedding lookup latency and the embedding copy latency. This is because of the
high-bandwidth \tdimm tensor operations and the fast
communication links utilized for moving the embeddings to GPU memory.
\fig{fig:perf_overall} summarizes the normalized performance of the five
design points across different batch sizes.
While \cpuonly occasionally achieves better performance than \cpugpu for low batch inference scenarios,
			the oracular \gpuonly consistently performs best on average,
			highlighting the advantages of DNN acceleration
using GPUs (and NPUs).
\tdimm  achieves an
average $84\%$ (no less than $75\%$) of the performance of such
unbuildable oracular GPU, demonstrating its performance merits and robustness
compared to \cpuonly and \cpugpu (an average $6.2\times$ and $8.9\times$ speedup).

\begin{figure}[t!] \centering
\includegraphics[width=0.48\textwidth]{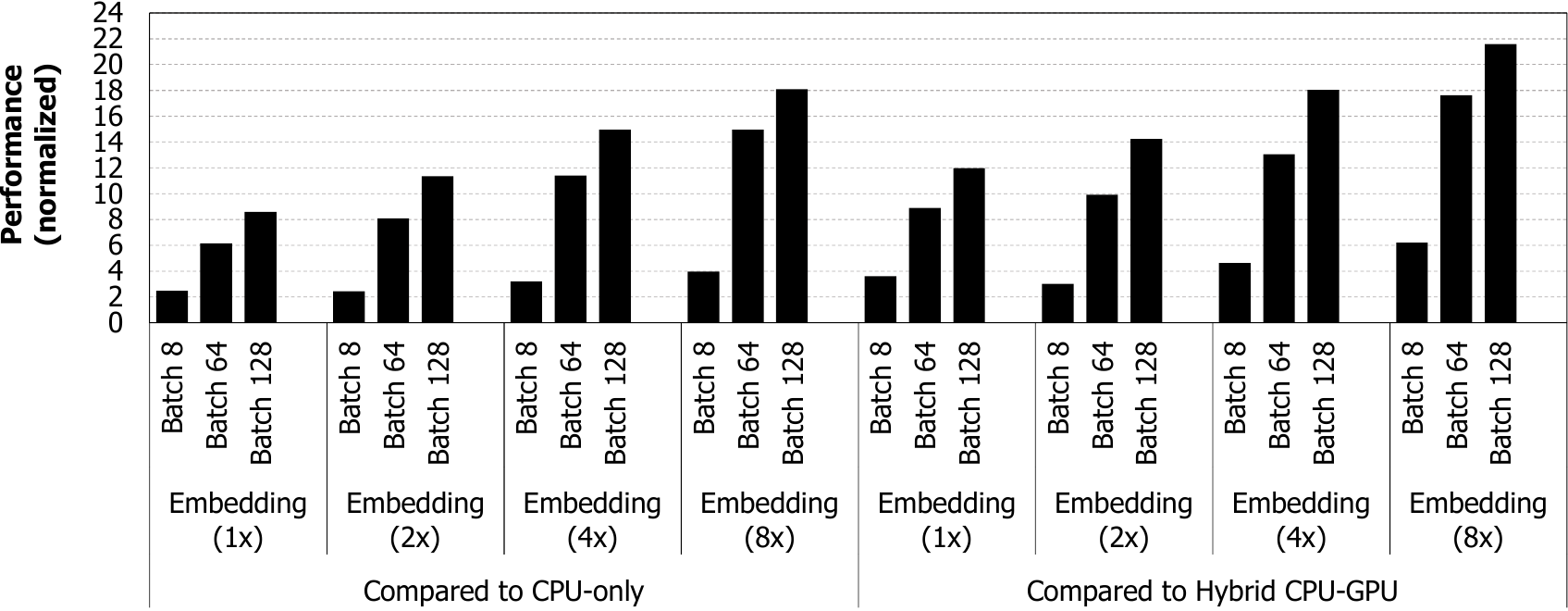}
\caption{
\tdimm performance with larger embedding size (up to 8$\times$).
Results are averaged across the four studied benchmarks.
}
\vspace{-0.5em}
\label{fig:sensitivity_larger_embedding}
\end{figure}

\begin{figure}[t!] \centering
\includegraphics[width=0.48\textwidth]{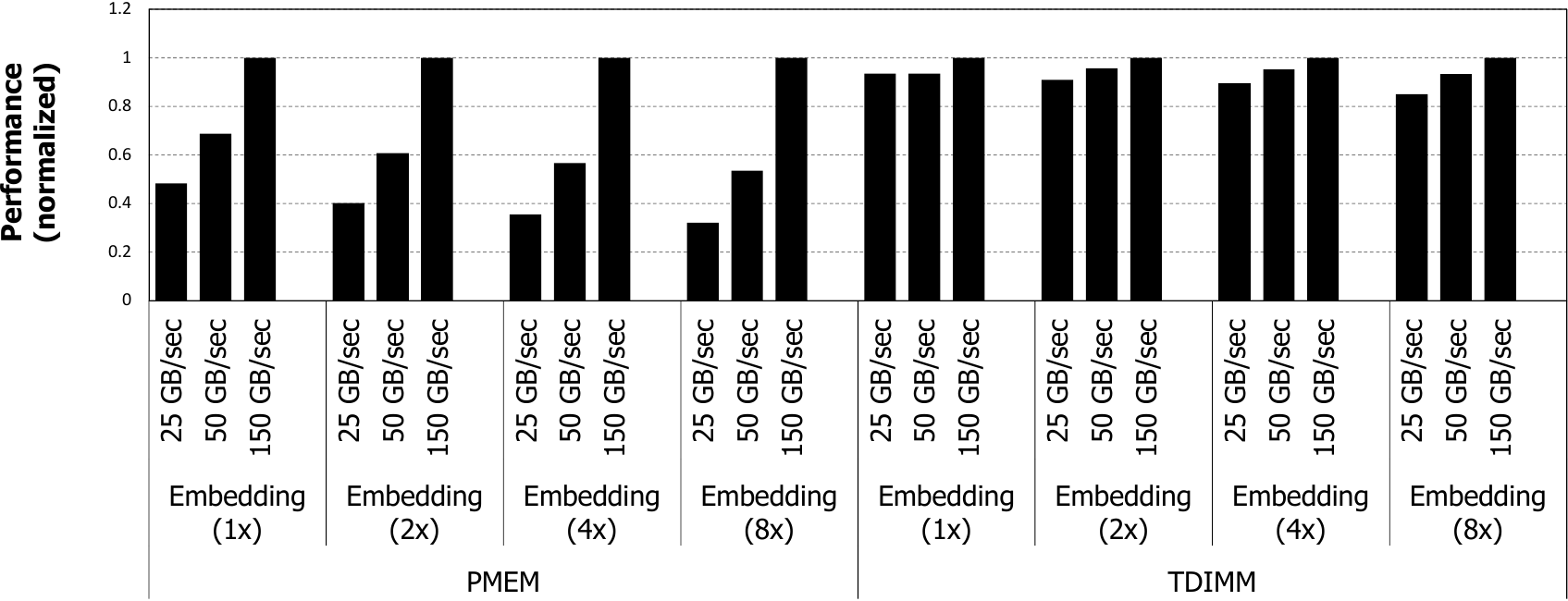}
\caption{
Performance sensitivity of \pmem (i.e., pooled memory without NMP acceleration) and \tdimm
	to the communication bandwidth.
	Results are averaged across the four studied benchmarks, and are normalized
	to the default configuration of 150 GB/sec data point.
}
\vspace{-0.5em}
\label{fig:sensitivity_comm_bw}
\end{figure}

\subsection{TensorDIMM with Large Embeddings}
\label{sect:sensitivity_large_embedding}

A key motivation of our work is to
provide a scalable memory system for embeddings and its tensor
operations. So far, we have assumed the default embedding size of each
workloads as summarized in \tab{tab:benchmarks}.  With the availablity of our
pooled memory architecture, DL practitioners can provision much larger
embeddings to develop recommender systems with superior model
quality. Because our evaluation is conducted over an emulated version of
\tdimms, we are not able to demonstrate the model quality
improvements larger embedding will bring about as we cannot train
these scaled-up algorithms (i.e., the memory capacity is still constrained by the GPU memory size).
Nonetheless, we conduct a sensitivity study
of the \tdimms for scaled-up embedding sizes which is shown in
\fig{fig:sensitivity_larger_embedding}. With larger embeddings,
	the embedding layers cause a much more serious performance bottleneck.
	\tnode shows even higher performance benefits under these settings, achieving
	an average $6.2$$-$$15.0\times$ and $8.9$$-$$17.6\times$ (maximum $35\times$)
	performance improvement than \cpuonly and \cpugpu, respectively.

\subsection{TensorDIMM with Low-bandwidth System Interconnects}
\label{sect:sensitivity_comm_bw}

To highlight the maximum potential of \tdimm, we discussed its merits under the
context of a high-bandwidth GPU-side interconnect.  Nonetheless, it is still
possible for \tdimm to be utilized under conventional, CPU-centric
disaggregated memory systems. Concretely, one can envision a system that
contains a pooled memory like \tnode interfaced	over a low-bandwidth system
interconnect (e.g., PCIe).  Such design point looks similar to the hybrid
\cpugpu except for one key distinction: the tensor operations are done
using the NMP cores but the reduced tensor must be copied to the GPU over a
slower communication channel.  \fig{fig:sensitivity_comm_bw} summarizes the
sensitivity of our proposal on the \texttt{TensorNode}$\leftrightarrow$GPU
communication bandwidth. We study both \pmem (i.e., disaggregated memory
		``without'' \tdimms) and \tdm under low-bandwidth system interconnects to
highlight the robustness our \tdimm brings about. Overall, \pmem is much
more sensitive to the communication bandwidth than \tdm (maximum $68\%$ performance loss)
because the benefits of NMP reduction is lost with \pmem. Our \tdimm design (\tdm) on the other
hand only experiences up to $15\%$ performance loss (average $10\%$) even with a $6\times$
lower communication bandwidth.  These results
highlight the robustness and the wide applicability of \tdimm.

\subsection{Design Overheads}
\label{sect:impl_overhead}

\tdimm requires no changes to the DRAM chip itself but adds a lightweight NMP
core inside the buffer device (\sect{sect:tensor_dimm}). We implement and
synthesized the major components of our NMP core on a Xilinx Virtex UltraScale+
FPGA board using Verilog HDL. We confirm that the added area/power
overheads of our NMP core is mostly negligible as it is dominated by the small
($1.5$ KB) SRAM queues and the $16$-wide vector ALU (\tab{tab:fpga}). From a
system-level perspective, our work utilizes GPUs as-is, so the major overhead
comes from the disaggregated \tnode design. Assuming a single \tdimm uses $128$
GB load-reduced DIMM~\cite{lrdimm}, its power consumption becomes $13$W when
estimated using Micron's DDR4 system power
calculator~\cite{micron_dram_power_estimator}.  For \tnode with
$32$ \tdimms, this amounts to a power overhead of ($13$$\times$$32$)=$416$W.  Recent
specifications for accelerator interconnection fabric endpoints (e.g., Open
		Compute Project~\cite{ocp}'s open accelerator module~\cite{zion}) employ a
TDP of $350$$-$$700$W so the power overheads of \tnode is expected to be
acceptable.

\section{Related Work}
\label{sect:discussion}

Disaggregated memory~\cite{disagg_mem_1,disagg_mem_2} is typically deployed as
a remote memory pool connected over PCIe, which helps increase CPU accessible
memory capacity. Prior work~\cite{mcdla,mcdla:cal,buddy_compression} proposed
system-level solutions that embrace the idea of memory disaggregation within
the high-bandwidth GPU interconnect, which bears similarity to our proposition
on \tnode. However, the focus of these two prior work is on DL training whereas
our study primarily focuses on DL inference.  Similar to our \tdimm, several
prior work~\cite{centaur,nda1,chameleon,mcn} explored the possibility of
utilizing the DIMM buffer device space to add custom acceleration logics. 
Alian et al.~\cite{mcn} for instance adds a lightweight CPU core inside the buffer
device to construct a memory channel network (MCN), seeking to amplify effective memory
bandwidth exposed to the overall system. While TensorDIMM's near-memory processing approach bears some similarity with 
MCN, direct
application of this design for embedding layer's tensor operations will likely lead
to sub-optimal performance because of MCN's inability to maximally utilize
DRAM bandwidth for embeddings. Concretely, Gupta et al.~\cite{dlrm:arch}
showed the performance of embedding lookup operations with
CPUs, reporting that the irregular, sparse memory access nature of
embedding operations render the CPU cache hit-rate to become extremely low. As
such, the latency to traverse the cache hierarchy leads to less than $5\%$ of the
maximum DRAM bandwidth being utilized, as opposed to the TensorDIMM NMP cores
being able to reach close to maximum DRAM bandwidth on average.
In general, the
scope of all these prior studies is significantly different from what our work focuses on.  To
the best of our knowledge, our work is the first in both industry and academia to identify and address the
memory capacity and bandwidth challenges of embedding layers, which several
hyperscalars~\cite{park:2018:fb,hestness:2019:ppopp,dean:2018:goldenage,facebook_dlrm,ocp_speech} deem as one of
the most crucial challenge in emerging DL workloads. 

Other than these closely related studies, a large body of prior work has
explored the design of  a single GPU/NPU architecture for DL~\cite{diannao,dadiannao,shidiannao,pudiannao,du:2015:micro,minerva,dnn_pim_reram,eyeriss,cambricon,isacc,neurocube,tabla,dnnweaver,intel:2017:fpl,gao:2017:tetris,intel:2018:fpga,rhu:2016:vdnn,jang:2019:mnnfast}
with recent interest on leveraging sparsity for further energy-efficiency
improvements~\cite{song:2015:eie,cnvlutin,cambriconx,stripes,bitpragmatic,intel:2017:icassp,intel:2017:fpga,whatmough:2017:isscc,whatmough:2017:hotchips,scnn,bittactical,rhu:2018:cdma}.
A scale-out acceleration platform for training DL algorithms was proposed by
Park et al.~\cite{cosmic} and a network-centric DL training platform has been
proposed by Li et al.~\cite{li:2018:ncdl}.  These prior studies are orthogonal
to our proposal and can be adopted  further for additional enhancements.

\section {Conclusion}
\label{sect:conclusion}

In this paper, we propose a vertically integrated, hardware/software co-design
that addresses the memory (capacity and bandwidth) wall problem of embedding
layers, an important building block for emerging DL applications. Our \tdimm
architecture synergistically combines NMP cores with commodity DRAM devices to
accelerate DL tensor operations.  Built on top of a disaggregated memory pool,
					 \tdimm provides memory capacity and bandwidth scaling for
					 embeddings, achieving an average $6.2$$-$$15.0\times$ and
					 $8.9$$-$$17.6\times$ performance improvement than conventional
					 \cpuonly and hybrid \cpugpu implementations of recommender systems.
					 To the best of our knowledge, \tdimm is the first that
					 quantitatively explores architectural solutions tailored for
					 embeddings and tensor operations.

\begin{table}[t!]
  \centering
	\vspace{0.8em}
  \caption{FPGA utilization of a single NMP core (i.e., SRAM queues and vector ALU with single-precision floating point (FPU) and fixed point ALU) on Xilinx Virtex UltraScale+ VCU1525 acceleration development board.}
\footnotesize
    \begin{tabular}{|ccccc|}
    \hline
	    & \textbf{LUT [\%]} & \textbf{FF [\%]} & \textbf{DSP [\%]} & \textbf{BRAM [\%]} \\
    \hline
    SRAM queues & 0.00 & 0.00 & 0.00 & 0.01 \\
    FPU					& 0.19 & 0.01 & 0.20 & 0.00 \\
    ALU					& 0.09 & 0.01 & 0.01 & 0.00 \\

    \hline
  \end{tabular}
\vspace{-0.5em}
  \label{tab:fpga}
\end{table}

\begin{acks}
  This research is supported by Samsung Research Funding Center of Samsung Electronics (SRFC-TB1703-03).
\end{acks}

\bibliographystyle{ACM-Reference-Format}
\bibliography{ref}

\end{document}